\documentclass{siamonline190516}

\usepackage{amsfonts}
\usepackage{graphicx}
\usepackage{overpic}
\usepackage{multirow}
\usepackage{subcaption}
\usepackage{mathtools}
\usepackage{psfrag}
\usepackage{dirtytalk}

\usepackage{epstopdf}
\usepackage{algorithmic}
\usepackage[shortlabels]{enumitem}
\usepackage{xspace}

\ifpdf
  \DeclareGraphicsExtensions{.eps,.pdf,.png,.jpg,.jpeg}
\else
  \DeclareGraphicsExtensions{.eps}
  \fi

\graphicspath{{./figures/}}

\usepackage{enumitem}
\setlist[enumerate]{leftmargin=.5in}
\setlist[itemize]{leftmargin=.5in}

\newsiamthm{claim}{Claim}
\newsiamremark{conjecture}{Conjecture}
\newsiamremark{remark}{Remark}
\newsiamremark{example}{Example}
\newsiamremark{hypothesis}{Hypothesis}
\newsiamremark{problem}{Problem}
\newsiamremark{assumption}{Assumption}

\usepackage{etoolbox}
\AtEndEnvironment{remark}{\null\hfill$\Diamond$}

\newcommand{\mcl}{\mathcal}

\newcommand{\mbf}{\mathbf}
\newcommand{\mb}{\mbf}
\newcommand{\mbb}{\mathbb}

\newcommand{\supp}{{\rm supp\ }}

\newcommand{\iidsim}{\stackrel{\text{iid}}{\sim}}

\newcommand{\dd}{{\rm d}}

\newcommand{\bv}{\mbf v}

\newcommand{\bx}{\mbf x}

\newcommand{\bb}{\mbf b}
\newcommand{\bW}{\mbf{W}}

\newcommand{\bd}{\mbf d}
\newcommand{\eps}{\epsilon}

\newcommand{\y}{y}

\renewcommand{\u}{u}

\newcommand{\N}{\mcl{N}}
\newcommand{\X}{\mcl{X}}
\newcommand{\Y}{\mcl{Y}}
\newcommand{\Z}{\mcl{Z}}
\newcommand{\A}{\mcl{A}}

\newcommand{\B}{\mcl{B}}

\newcommand{\U}{\mcl{U}}
\newcommand{\D}{\mcl{D}}
\newcommand{\R}{\mcl{R}}
\newcommand{\mP}{\mcl{P}}
\newcommand{\tU}{{\mcl{V}}}
\newcommand{\tY}{{\mcl{W}}}
\newcommand{\tZ}{{\mcl{S}}}
\newcommand{\tu}{v}
\newcommand{\ty}{w}
\newcommand{\tz}{s}

\newcommand{\Id}{{\rm Id}}

\newcommand{\T}{\mathsf{T}}

\newcommand{\mT}{\mcl{T}}
\newcommand{\tT}{\widetilde{\T}}
\newcommand{\mH}{\mcl{H}}
\renewcommand{\S}{\mathsf{S}}
\newcommand{\F}{\mathsf{F}}

\newcommand{\mF}{\mathcal{F}}

\newcommand{\G}{\mathsf{G}}

\newcommand{\Q}{\mathsf{Q}}
\newcommand{\mQ}{\mcl{Q}}
\newcommand{\J}{\mathcal{J}}

\newcommand{\PP}{\mbb{P}}

\newcommand{\RR}{\mbb{R}}
\newcommand{\EE}{\mbb{E}}
\newcommand{\NN}{\mbb{N}}

\newcommand{\WGANGP}{{\rm WGP}\xspace}
\newcommand{\LSGAN}{{\rm LS}\xspace}
\newcommand{\Law}{{\rm Law}\xspace}
\newcommand{\MGAN}{{M-GAN}\xspace}
\newcommand{\MdGAN}{{M-GAN}\xspace}
\newcommand{\MGANs}{{M-GAN}s\xspace}

\definecolor{darkred}{rgb}{.7,0,0}

\definecolor{darkgreen}{rgb}{.15,.55,0}

\definecolor{darkblue}{rgb}{0,0,0.7}

\usepackage{amsopn}

\DeclareMathOperator*{\minimize}{minimize}

\DeclareMathOperator*{\st}{s.t.}

\reversemarginpar
\usepackage[colorinlistoftodos,prependcaption,textsize=tiny]{todonotes}

\headers{Conditional Sampling with Monotone GANs}{R.\ Baptista, B.\ Hosseini, N.\ B.\ Kovachki, Y.\ Marzouk}

\title{Conditional Sampling with Monotone GANs: \\  from Generative Models to Likelihood-Free Inference}

\author{Ricardo Baptista\thanks{California Institute of Technology, Pasadena, CA 91106, USA
  (\email{rsb@caltech.edu})} 
  \and Bamdad Hosseini\thanks{University of Washington, Seattle, WA 98195, USA
  (\email{bamdadh@uw.edu})} 
  \and Nikola B.\ Kovachki\thanks{NVIDIA, Santa Clara, CA 95051, USA
  (\email{nkovachki@nvidia.com})} \and Youssef Marzouk\thanks{Massachusetts Institute of Technology, Cambridge, MA 02139, USA (\email{ymarz@mit.edu})} %
  }

\ifpdf
\hypersetup{
  pdftitle={Conditional Sampling with Measure Transport},
  pdfauthor={R. Baptista, B. Hosseini, N. B. Kovachki, Y. Marzouk}
}
\fi

\begin{document}

\maketitle

\begin{abstract}
We present a novel framework for conditional sampling of probability measures, using block triangular transport maps. We develop the theoretical foundations of block triangular transport in a Banach space setting, establishing general conditions under which conditional sampling can be achieved and drawing connections between monotone block triangular maps and optimal transport.
Based on this theory, we then introduce a computational approach, called monotone generative adversarial networks (\MGANs), to learn suitable block triangular maps. 
Our algorithm uses only samples from the underlying joint probability measure and is hence likelihood-free. 
Numerical experiments with \MGAN demonstrate accurate sampling of conditional measures in synthetic examples, Bayesian inverse problems involving ordinary and partial 
differential equations, and probabilistic image in-painting.
\end{abstract}



\section{Introduction}\label{sec:introduction}

Conditional simulation can be viewed as the process of generating samples from certain ``slices'' of a probability measure $\nu \in \PP(\U \times \Y)$. Intuitively, simulating $\u \in \U$ conditioned on a given value of $\y \in \Y$ amounts to restricting $\nu$ 
along a hyperplane $\y = \y^\ast$, renormalizing, and generating samples from the resulting distribution. Conditional sampling problems are ubiquitous in statistics, applied mathematics, and engineering, where $\u$ may represent an output or prediction of interest and $\y$ may represent a variable that is observed.\footnote{Our notation here is chosen to be consistent with the literature on Bayesian inverse problems, where $y$ denotes the data and $u$ denotes an unknown of interest.} 

Many \textit{supervised learning} algorithms such as ridge, lasso, or neural network regression assume a finite-dimensional parameterization of $u$ and use a statistical model of the observations, $\nu(\cdot \vert u)$, perhaps paired with some penalization scheme, to construct a point estimator $\hat{\u}(\y^\ast)$ of $\u$ for any $\y^\ast$. In the probabilistic setting described above, where the $\U$-marginal $\nu_\U$ is naturally interpreted as a prior measure, such point estimators may coincide with the mode of $\u$ conditioned on $\y^\ast$ under specific likelihood and prior models~\cite{hastie2009elements}. Fully Bayesian methods, however, go further and seek to characterize the entire conditional measure $\nu( \cdot \vert \y^\ast)$, thereby providing a natural way of quantifying uncertainty in the predicted outputs. Gaussian process regression is a canonical example, where $\u$ is an element of an infinite-dimensional space and $\nu$ is also Gaussian on the 
product space.

\textit{Inverse problems} in the Bayesian setting~\cite{kabanikhin,stuart-acta-numerica} fall into the aforementioned framework as well; here, one seeks to recover an unknown parameter $u$ from a realization of \textit{indirect} and noisy observations $\y^\ast$, where $\u$ is typically infinite-dimensional. A prototypical inverse problem takes the form
\begin{equation}\label{generic-inv-prob}
  \mcl L(\u) p = 0, \qquad \y =  g(p) + \epsilon, 
\end{equation}
where $\u \in \U$ represents the parameter of interest, $p \in \mP$ is a state variable, and $\mcl L(\u)$ is an operator acting on $p$, parameterized by $\u$. Here, $g\colon \mP \to \Y$ is an observation operator that extracts $\y \in \Y$ from $p$, and $\epsilon \in \Y$ is a random variable representing observational noise; $\U$, $\Y$, $\mP$ are assumed to be Banach spaces. For instance, $\mcl L$ could be a partial differential operator and $g$ could return pointwise evaluations of the PDE solution $p$, defined with appropriate boundary conditions (see \Cref{sec:darcy-flow} for a concrete example). 
From a probabilistic perspective, \eqref{generic-inv-prob} specifies the conditional distribution $\nu( \cdot \vert u )$. In the Bayesian setting \cite{stuart-acta-numerica}, one also endows $\u$ with a prior and thus fully specifies the joint probability measure 
$\nu$, with the goal of then characterizing the posterior measure $\nu( \cdot | \y^\ast)$.

The common challenge in the applications outlined above is therefore to sample from a conditional measure $\nu(\cdot | \y^\ast)$, as sampling enables the estimation of arbitrary moments or other expectations.  
Markov chain Monte Carlo (MCMC) algorithms are widely used for this purpose and provide asymptotically exact estimates, but require repeatedly evaluating the likelihood (e.g., solving~\eqref{generic-inv-prob} in the case of Bayesian inverse problems); moreover, one must simulate an entirely new Markov chain for each new value of $y^\ast$. Also, the performance of most MCMC algorithms is quite sensitive to the choice of prior and likelihood models \cite{stuart-mcmc, hairer2014spectral, hosseini2019two, hosseini2018spectral, vollmer2015dimension}. These issues often limit the utility of MCMC in large-scale applications.
Variational inference (VI) methods \cite{blei2017variational, fox2012tutorial, zhang2018advances} offer an alternative to MCMC by approximating the conditional measure $\nu( \cdot \vert \y^\ast)$
with a measure $\nu_\theta$ chosen from a certain tractable family parameterized by $\theta$. For example, one can take $\nu_\theta$ to be the family of Gaussian measures on $\U$ parameterized by their means and covariance operators.  While VI can be significantly more efficient than MCMC,
the accuracy of variational inference is very much limited by the quality of the approximating family (See \cite{blei2017variational} for a more detailed discussion  and for comparisons between MCMC and VI.)

In this article, we present and analyze a novel framework for conditional sampling using transportation of measure.
Our methods fall under the umbrella of VI, although the optimization problems and distributional approximations of interest to us are not standard in VI: our family of approximating measures $\nu_\theta$ comprises the pushforwards of a chosen reference measure by parameterized \emph{block-triangular transport maps}; also, we solve optimization problems whose objectives involve statistical divergences inspired by optimal transport (OT) distances, rather than the Kullback--Leibler (KL) divergence. In this light, our methods are closely related to modern generative models in machine learning (ML), such as generative adversarial networks (GANs) \cite{goodfellowGANS} and normalizing flows \cite{kobyzev2019normalizing}. Another feature differentiating our approach from MCMC and standard VI is the ability to approximate the entire family of conditionals $\nu(\cdot | y)$ by solving a single optimization problem, making it attractive for settings where conditional simulation for a large collection of observation values is desired. A further distinguishing feature is that our approach is entirely data-driven: the approximate conditionals $\nu_\theta(\cdot | y^\ast)$ are computed only using samples from the joint measure $\nu$.

In the remainder of this section, we give a summary of our main contributions, followed by a review of relevant literature.

\subsection{Main contributions}\label{subsec:main-contributions}

Consider a {\it reference measure} $\eta$ and a {\it target measure} $\nu$, both of which are Borel measures on the separable Banach space $\Y \times \U$. We assume that $\eta$ is known and can be simulated
at low cost; for example, we can choose $\eta$ to be the standard Gaussian measure whenever $\Y$, $\U$ are finite-dimensional, or an appropriate Gaussian process in the Banach space setting.
Our goal is to generate approximate samples from $\nu( \cdot \vert y^\ast)$.
To this end, we  pose 
 optimization problems of
 the form 
 \begin{equation}\label{opt-prob-intro}
   \left\{
     \begin{aligned}
       &\min_{\F, \G} && \D(\T_\sharp \eta, \nu) + \R(\T), \\
       &\st && \T(y, u) = \big( \F(y), \G(\F(y), u) \big), \\
       &    && \F: \Y \to \Y,\quad \G: \Y \times \U \to \U,
     \end{aligned}
     \right.
   \end{equation}
   where 
   $\D$
   is a  statistical divergence  on  $\PP(\Y \times \U)$,  $\R$ 
   is an appropriate regularization term,  
   and $\F$, $\G$ are parametric maps.
   We make three main contributions in this work:

   \begin{itemize}
   \item We present a theoretical analysis of \eqref{opt-prob-intro} in an idealized setting
     where $\U$, $\Y$ are Banach spaces and $\R \equiv 0$. Our analysis yields three primary results:
     (a)  If $\T_\sharp \eta = \nu$, then 
     $\G(y^\ast, \cdot)_\sharp \eta_\U
     = \nu( \cdot | y^\ast)$, where $\eta_\U$ is the $\U$-marginal of $\eta$;
     (b) Under very general conditions on $\eta$, $\nu$ and
     for wide choices of $\D$, problem~\eqref{opt-prob-intro} has a minimizer $\T^\dagger$
     that satisfies $\T^\dagger_\sharp \eta = \nu$; (c) Under appropriate monotonicity
     constraints on $\T$, and when $\Y$ and $\U$ are finite-dimensional Euclidean spaces, the resulting conditioning map $\G^\dagger(y^\ast, \cdot)$ is also unique and is the solution to an OT problem (in fact it is a conditional Brenier map \cite{carlier2016vector}).
     We present these results in \Cref{sec:theory}.

   \item Motivated by this theoretical foundation, we present a computational framework called
     {\it Monotone Generative Adversarial Networks} (\MGANs) that approximates \eqref{opt-prob-intro}
     in three steps: (a) Take $\D$
     to be  an approximate Wasserstein-1 type distance; (b) parameterize $\F$, $\G$ as
     neural networks; (c) impose monotonicity on $\T$ via the regularization term $\R$. We then solve the resulting optimization problem using stochastic gradient descent (SGD)
     to obtain a minimizer $\G^\dagger$. Given a $y^\ast$, we then draw samples $u_j \iidsim \eta_\U$
     and evaluate $\G^\dagger(y^\ast, u_j)$ to obtain approximate samples from $\nu( \cdot | y^\ast)$.
     The \MGAN framework is outlined in \Cref{sec:MGAN}.

   \item We evaluate the performance of the \MGAN approach numerically, with experimental settings ranging from low-dimensional synthetic problems to high-dimensional ML applications and infinite-dimensional Bayesian inverse problems involving PDEs. These experiments can be found in \Cref{sec:numerics}.
   \end{itemize}

A core feature of the \MGAN framework is that to solve \eqref{opt-prob-intro} numerically, we only require samples from the joint measure $\nu$, yet the map $\G^\dagger$ characterizes all of the conditionals  $\nu( \cdot | y)$. In other words, \MGAN is entirely data-driven and does not require evaluations of a likelihood function or prior density; more generally, it does not require any explicit knowledge or modeling assumptions on the relationship between $u$ and $y$. Moreover, since the computed $\G^\dagger$ can be evaluated at multiple values of $y^\ast$ without any additional optimization, the cost of inference is ``amortized'' over $\y^\ast$ 
\cite{gershman2014amortized, rezende2015variational, zhang2018advances, cranmer2020frontier, lueckmann2021benchmarking}.

\subsection{Relevant literature}\label{subsec:relevant-literature}
Conditional sampling is an active area of research in computational statistics, ML, and applied mathematics. Conventional methods such as MCMC and VI, as mentioned earlier, have a rich and active literature but a thorough review of these topics is outside the present scope.  Instead we focus on literature pertaining to conditional sampling and transportation of measure.

\subsubsection{Measure transport in uncertainty quantification}
 The use of transport maps for conditional sampling has been  
 explored in the uncertainty quantification (UQ) and inverse problems
 communities~\cite{marzouk2016sampling,marzouk-opt-map,spantini2019coupling, siahkoohi2021preconditioned}.
 For problems in Bayesian inference and ML~\cite{bigoni2019greedy, jaini2019sum, papamakarios2017MAF}, a common approach is to seek monotone triangular maps that approximate the classic Knothe--Rosenblatt (KR) rearrangement~\cite{santambrogio2015optimal}. 
By construction, components of the KR rearrangement push forward a product reference measure $\eta$ to the target conditionals, which is precisely what is desired for conditional sampling. 
 While the KR rearrangement can be written explicitly in terms of marginal-conditional distribution and quantile functions, direct computation using this definition is typically infeasible. The approach of \cite{marzouk2016sampling,marzouk-opt-map,spantini2019coupling} instead is to formulate problems akin to \eqref{opt-prob-intro} by choosing $\D$ to be the
 Kullback--Leibler (KL) divergence, taking the reference $\eta$ to be the standard Gaussian, and parameterizing $\T$ in a space of monotone triangular functions. These choices naturally affect the accuracy of the resulting transport. Also, most triangular map representations (with the exception of \cite{zech2022sparseB}) are limited to finite-dimensional input spaces $\U$ and $\Y$, and the fully triangular form of $\T$ requires selecting a particular ordering of the coordinate bases. We demonstrate in \Cref{sec:banana} that this choice can have a significant impact on accuracy in practice. Furthermore, monotone parameterizations of $\T$ can lead to poorly behaved optimization problems (e.g., with many local minima) unless one exercises sufficient care, as described in~\cite{baptista2020adaptive}.
Our \MGAN framework addresses the aforementioned drawbacks of triangular transport maps by generalizing the formulation of \cite{marzouk2016sampling,marzouk-opt-map,spantini2019coupling} in several ways: (a) we allow wider choices of $\D$; (b) we ask only for $\T$ to be block triangular, such that no ordering of coordinate bases for $\U$ or $\Y$ is needed; and (c) we establish validity of our formulation on infinite-dimensional Banach spaces. To achieve conditional sampling, we do not even require $\T$ to be monotone, although we do impose monotonicity in practice and enforce it in some of our theoretical results (e.g., in making a link to OT).

Analysis of triangular transport maps is a classical topic going back to the works of Knothe \cite{knothe1957contributions} and Rosenblatt \cite{rosenblatt1952remarks}. Basic properties of such maps, such as existence, uniqueness, and regularity, have since been studied in general settings including infinite-dimensional Banach spaces \cite{bogachev2005triangular, bogachev2006nonlinear}. 
Applied analysis of triangular maps,  pertaining to algorithms, has become of interest much more recently: 
\cite{zech2022sparse,zech2022sparseB} show that under appropriate assumptions on the reference and target measures, the KR rearrangement is analytic on the finite- or infinite dimensional hypercube and can be well approximated with sparse polynomials or deep ReLU networks;
\cite{irons2021triangular} considers a variational characterization of the KR rearrangement akin to \eqref{opt-prob-intro} and studies the statistical consistency and convergence of the empirical approximation of the map given samples from the target $\nu$;
\cite{wang2022minimax} establishes optimal minimax rates of convergence for nonparametric density estimators based on triangular and other transport maps, by adapting techniques from $M$-estimation and empirical process theory to the transport setting;
 \cite{jaini2020tails} analyzes the tail behavior of triangular maps, revealing an intricate balance between the tails of the reference and target measures and the expressive power of Lipschitz maps;
 and \cite{cui2023scalable} presents an efficient and scalable tensor train parameterization of 
 triangular maps for conditional sampling.

Our theoretical contributions are distinct from the aforementioned articles in four aspects:  (a)
we do not limit ourselves to triangular/KR maps and instead consider the much more  general problem in~\eqref{opt-prob-intro}; (b) we allow for a generic choice of $\D$ as opposed to the KL divergence; (c) we develop existence and convergence results for the conditioning map $\G$ as opposed to the full map $\T$; and (d) we connect our block triangular construction to recent results in OT.

\subsubsection{Measure transport in ML}
Measure transport problems have also attracted considerable interest in the ML literature, particularly for generative modelling \cite{ng2002discriminative, jebara2012machine}.
Following \cite{jebara2012machine}, we say that a ML model or algorithm is ``generative''
if it characterizes the joint measure $\nu$ rather than the conditional measure $\nu( \cdot | \y^\ast)$. In this definition, the map $\T$ obtained by solving \eqref{opt-prob-intro} is a generative model. Popular generative models of relevance to our \MGAN framework are
GANs \cite{goodfellowGANS, goodfellow2016nips},
normalizing flows \cite{rezende2015variational, papamakarios2019normalizing, kobyzev2019normalizing},
and, to some extent, variational autoencoders (VAEs) \cite{kingma-VAE, doersch2016tutorial}.
All of these methods and their variants solve problems of a form similar to \eqref{opt-prob-intro}, but with three core differences: 
(a) the reference measure $\eta$ in GANs and VAEs is often defined on a lower-dimensional space, enabling natural dimension reduction; 
(b) the map $\T$ is parameterized directly and the maps  $\F, \G$ are omitted from the formulation; and
(c) the regularization term $R$ is not identified, or its impact is not analyzed explicitly.
In GANs, the map $\T$ is often parameterized by a single neural network and $\D$ is taken to be a GAN loss function, which can be viewed as an approximation to a Wasserstein-type
distance or a variational form of a statistical divergence \cite{nowozin2016f, wGAN}. 
Normalizing flows represent $\T$ as a \emph{composition} of invertible and often \emph{triangular} \cite{jaini2019sum} neural networks, typically interleaved with permutations, and may choose $\D$ to be the KL divergence \cite{papamakarios2019normalizing}.
In this light, NFs are closely related to the triangular maps of \cite{marzouk2016sampling,marzouk-opt-map,spantini2019coupling, siahkoohi2021preconditioned}\footnote{In fact, this connection to triangular maps and our analysis in \Cref{sec:theory} imply that normalizing flows can easily be retrofitted for conditional sampling, simply by appending the conditioning variables $\y$ as additional inputs and constraining the permutation layers appropriately \cite{cranmer2020frontier, radev2020bayesflow, ardizzone2019guided, marzouk2016sampling}.}. 
VAEs pose a slightly different problem to GANs and normalizing flows by parameterizing both $\T$ and its inverse $\T^{-1}$ as separate neural networks and approximating them simultaneously. The KL divergence is again employed in most VAE applications.

As the name \MGAN suggests, our proposed framework is closely related to  GANs. In fact,
one can view \MGAN as a combination of GANs and normalizing flows with a particular parameterization of the
map $\T$. However, we emphasize that the aforementioned generative models aim to approximate the map $\T$, with the ultimate goal of sampling the joint distribution $\nu$. The task of conditioning $\nu$ is not of direct interest, and is often tackled in a secondary step using Bayes' rule or other standard (or ad hoc) conditioning techniques. Thus, a defining feature of \MGAN is that it allows us to directly characterize the family of conditionals $\nu( \cdot | y)$ through the map $\G$. We then obtain the generative model $\T$ as a by-product.

Several previous efforts to adapt generative models for conditional sampling exist in the ML literature.   
Most notably, \cite{mirza2014conditional, ivanov2018variational} define conditional GANs and VAEs by
training neural networks that depend on the joint variables $(\y, \u)$ to obtain maps that can
generate samples from multi-modal distributions. However, these formulations are limited to settings
where $\y$ is a discrete variable and many $\u$ samples are available for a given $\y$; such data is typically 
not available for continuous $\y$. More recently, \cite{ivanov2018variational, belghazi2019learning}
addressed the more difficult problem of approximating all the conditionals of the
joint distribution $\nu$ by employing a weighted loss function
over all possible choices of conditionals. This approach has two major drawbacks: the
loss does not guarantee that any particular conditional is obtained correctly, and
the problem quickly becomes infeasible in high- or infinite-dimensional settings. 
 \cite{lindgren2020conditional} considered the problem of correctly extracting
a single conditional from a fixed generative model using a variational inference loss. The proposed method 
must be re-trained for \emph{each} new value of $\y$, in contrast to \MGAN where the map $\G^\dagger$
characterizes the entire conditional family $\nu( \cdot |y)$ simultaneously.
Finally, probabilistic diffusion models have recently been adapted for the solution of 
inverse problems~\cite{song2021solving, batzolis2021conditional, song2023pseudoinverse}. This is performed by either %
learning a problem-agnostic diffusion model given prior samples and appropriately %
"guiding" the reverse process to generate samples from 
the conditional of interest using the measurement model, or by learning a model for specific problems, such as super-resolution, given samples from the joint distribution of parameters and observations and fixing an observation for conditional sampling during the reverse process. The latter approach for amortized training and sampling is analogous to the one presented in \Cref{subsec:main-contributions}. Diffusion models promise great flexibility for 
solving inverse problems involving high-resolution images, %
albeit at the high cost of training and inverting a large diffusion model.

The recent articles \cite{adler2018deep, ray2022efficacy, zhou2022deep} are closest to our construction.
The approaches of \cite{adler2018deep, ray2022efficacy} are similar to each other and can be viewed as particular versions of \MGAN by taking $\F$ to be the identity map, omitting the monotonicity penalty/constraint on $\T$, and choosing a particular form of $\D$ and $\G$. 
The article \cite{zhou2022deep} considers a similar situation by assuming $\eta$ to be Gaussian, choosing $\D$ to be an $f$-divergence, and parameterizing $\G$ with a neural network. The theoretical exposition in \cite{zhou2022deep} is well aligned with our results in \Cref{sec:theory}, 
although they only consider the case where $\Y, \U$ are finite dimensional Euclidean spaces
and do not make the connection 
to OT.
Our work can be differentiated from these efforts in three directions: 
(a) our theoretical results and algorithms are valid on infinite-dimensional Banach spaces, a setting that is crucial for PDE inverse problems;
(b) our formulation is more general, placing minimal assumptions on $\D$, the parameterization of $\T$, or the choice of reference distribution $\eta$; 
(c) by including a monotonicity penalty, we are able to provide further understanding of the solution of~\eqref{opt-prob-intro} by connecting our minimizer to OT.

We briefly mention another relevant body of work that directly estimates (and selects) models for conditional densities. Examples include mixture models parameterized by neural networks~\cite{bishop1994mixture, rothfuss2019conditional}, or more flexible nonparametric models~\cite{arbel18a, shiga2015direct, ambrogioni2017kernel}. However, parametric methods impose structural constraints on the target conditional densities and do not focus on conditional \emph{sampling} as we do here, while nonparametric (e.g, kernel) methods typically have growing sampling costs with the size of the dataset and require careful 
regularization.

\subsubsection{Connection to OT}
Given two measures $\eta, \nu \in \PP(\Z)$, the Monge problem of optimal transportation seeks maps
$\T$ that satisfy $\T_\sharp \eta = \nu$ while minimizing
functionals of the form $\int c( z, \T(z) ) \eta(\dd z)$, for appropriate cost functions
$c: \Z \times \Z \to \R$. 
Our analysis and the construction of the \MGAN framework are strongly inspired by OT. In fact, existence and uniqueness results from OT can be extended to minimizers of \eqref{opt-prob-intro}. Furthermore, the monotonicity penalty in the \MGAN framework is directly motivated by
uniqueness results for the well-known Brenier maps \cite{mccann1995existence} and
their block triangular extensions \cite{carlier2016vector}. 
The articles \cite{carlier2010knothe, muzellec2019subspace} are also closely related to our 
(block) triangular constructions.
We present a more detailed discussion
of how our approach relates to OT in \Cref{subsec:monotonicity-OT}.

We note, however, that there are fundamental differences between problem \eqref{opt-prob-intro} and the
OT problem. Most importantly, OT maps are constrained to push
the reference $\eta$ to the target $\nu$ exactly while minimizing a transport cost; instead, we ask only to minimize $\D( \T_\sharp \eta, \nu)$ and thus may not match the target measure $\nu$ exactly. Furthermore, we restrict the function space to which $\T$ belongs and regularize this map via the penalty
$R$. These relaxations allow us to obtain ``nicer'' transport maps that can be computed in a stable
manner. Despite these relaxations, we demonstrate in \Cref{sec:numerics_OT} that
\MGAN maps converge to certain OT maps if $D$ and the space in which one seeks $\T$ are chosen correctly.
Our results in this direction draw on results from \cite{carlier2016vector}. 
This observation suggests that \MGAN can serve as a numerical method for 
approximating (conditional) OT maps, which is a topic that has attracted much interest in the 
ML community \cite{genevay2016stochastic, seguy2018large, korotin2021neural}. We also mention recent articles \cite{kan2022multivariate, taghvaei2022optimal, al2023optimal} that use conditional OT strategies resembling M-GANs for filtering and data assimilation.

\subsection{Outline}
The remainder of this paper is organized as follows. \Cref{sec:theory} establishes the necessary conditions for performing conditional sampling via block triangular transport maps, and discusses the existence and uniqueness of these maps in relation to those found via OT.  \Cref{sec:MGAN} presents our framework for monotone transport map approximation.  
\Cref{sec:numerics} presents numerical results for generative modeling and the solution of Bayesian inverse problems, followed by a concluding discussion in \Cref{sec:conclusions}.

\section{Theoretical foundations}\label{sec:theory}

In this section, we develop a theoretical analysis of
problem~\eqref{opt-prob-intro} in idealized settings, which serves as a foundation
for the \MGAN framework introduced in Section~\ref{sec:MGAN}.

Let $\U$, $\tU$, $\Y$, $\tY$ be separable Banach spaces
with  Borel $\sigma$-algebras  $\B(\U)$, $\B(\tU)$, $\B(\Y)$, $\B(\tY)$, respectively.
Define the product spaces
$\Z \coloneqq  \Y \times \U$ and $\tZ \coloneqq  \tY \times \tU$, with 
corresponding product  Borel $\sigma$-algebras $\B(\Z)$ and $\B(\tZ)$.
Let $\PP(\U)$,  $\PP(\Y)$, $\PP(\Z)$, $\PP(\tU)$, $\PP(\tY)$, $\PP(\tZ)$ denote spaces
of Borel probability measures on their respective Banach spaces.
For a measure
 $\mu \in \PP(\Z)$ (resp.\ $\in \PP(\tZ)$) we use $\mu_\U$ and $\mu_\Y$
 (resp.\ $\mu_{\tU}$ and $\mu_\tY$) to denote the marginals of $\mu$ on $\U$ and $\Y$ (resp.\ $\tU$ and $\tY$).  Finally, for any set
 $B \in \Z $ we define the slice  
 $B_y \coloneqq  \{ u :  (u,y) \in B\}$.
 In what follows, $\U$ will represent the parameter space and $\Y$ will represent the space of data on which we condition, with $\tU$ and $\tY$ being their corresponding ``reference'' spaces. 
We now recall the definition of regular conditional measures:
 \begin{definition}[Regular conditional measures]\label{def:conditional-measure}
   Let $\mu \in \PP(\Z)$. We say $\mu( \cdot | y)$ is a {\it system of regular conditional measures}
   for $\mu$ if:
  \begin{enumerate}
  \item $\forall y \in \Y$, $\mu( \cdot | y)$ is a probability measure on $\B(\U)$. 
  \item $\forall A \in \B(\U)$, the function $y \mapsto \mu(A | y)$ is measurable with
    respect to $\B(\Y)$ and is $\mu_\Y$-integrable.
  \item $ \forall B \in \B(\Z)$, it holds that
      $\mu( B) = \int_\Y \mu(B_y | y) \mu_\Y( \dd y)$.
  \end{enumerate}
\end{definition}

In what follows, we often refer to systems of regular conditional measures simply as
systems of conditional measures or ``conditionals.''
 By \cite[Cor.~10.4.15]{bogachev2} we have the following existence and uniqueness result: 

 \begin{proposition}\label{prop:product-conditional-measures}
   Consider the above setting with $\mu \in \PP(\Z)$. Then it holds that 
  \begin{enumerate}[(a)]
  \item (Existence) There exist Radon conditional measures $\mu( \cdot | y)$ of $\mu$
  for all $y \in \Y$. 
  \item (Uniqueness) The conditional measures $\mu(\cdot | y)$ are unique
    up to $\mu_\Y$ null sets.
  \end{enumerate}
\end{proposition}

Now consider a {\it reference measure} $\eta = \eta_\tY \otimes \eta_\tU \in \PP(\tZ)$, that is,
of product form, 
and a {\it target measure} $\nu \in \PP(\Z)$. Our goal
throughout this article is to characterize the conditionals $\nu( \cdot | y)$ via a transformation
of the reference measure $\eta$---specifically, a transformation of the marginal $\eta_\tY$. To this end,
consider a {\it block triangular map} of the form 
\begin{equation}
  \label{transport-map}
  \T: \tZ \to \Z, \qquad \T(\ty, \tu)  = \big ( \F(\ty), \G( \F(\ty), \tu) \big ),
\end{equation}
which in turn is defined through  the maps
\begin{equation}\label{F-G-map-definition}
  \F: \tY \to \Y, \qquad \G: \Y \times \tU \to \U. 
\end{equation}

\begin{remark}
  We refer to $\T$ as a \emph{block triangular map} since, in the  setting where $\tU$, $\U$, $\tY$, and $\Y$
  are finite-dimensional Euclidean spaces, the Jacobian matrix of $\T$ is block triangular.
We note that such maps are also simply called triangular in the literature;
see for example \cite[Sec.~10.10(vii)]{bogachev2}. However, we prefer the term block triangular
to set our parameterizations apart from strictly triangular maps such as the KR rearrangement considered in
\cite{marzouk2016sampling} or the elementary maps in normalizing flows \cite{kobyzev2019normalizing, papamakarios2019normalizing}.
We demonstrate in \Cref{sec:banana} that block triangular maps perform quite differently
from strictly triangular maps in practice.
\end{remark}

\subsection{Block triangular transport}
\label{subsec:block-triangular-transport}

The following theorem is the cornerstone of our methodology for
approximating the conditionals of $\nu$ via block triangular transport.

\begin{theorem}\label{thm:triangular-maps-condition}
  Consider a reference $\eta = \eta_\tY \otimes \eta_\tU \in \PP(\tZ)$ and a target $\nu \in \PP(\Z)$ and  
  let $\T$ be a {block} triangular map of the form \eqref{transport-map} satisfying
   $ \T_\sharp \eta = \nu$. Then for $\F_\sharp \eta_\tY$-a.e.\thinspace $y$ it holds that
   $ \G(y, \cdot)_\sharp \eta_\tU = \nu( \cdot | y).$
\end{theorem}

\begin{proof}
  Consider the maps $ \tT\colon (y, \tu) \mapsto (y, \G(y, \tu))$ and
  $(\F \times \Id)\colon (\ty,\tu) \mapsto (\F(\ty), \tu)$ and observe that
  $\T = \tT \circ (\F \times \Id)$.
  Let $B  \in \B(\Z)$. We have by  the
  hypothesis of the theorem that 
  \begin{equation*}
    \begin{aligned}
      \int_B \nu( \dd z)
      & = \int_B \T_\sharp \eta (\dd z)  = \int_{\tT^{-1}(B)}  (\F \times \Id)_\sharp \big(\eta_\tY\otimes \eta_\tU \big)( \dd \ty, \dd \tu) \\
      &= \int_{\tT^{-1}(B)}  \big( \F_\sharp \eta_\tY\otimes \eta_\tU \big)( \dd y, \dd \tu)
      = \int_{\tT^{-1}(B)}  \big( \nu_\Y \otimes \eta_\tU \big)( \dd y, \dd \tu),
    \end{aligned}
  \end{equation*}
  where the last identity follows from the observation that $\T_\sharp \eta = \nu$ implies
  $\F_\sharp \eta_\tY = \nu_\Y$ due to the product structure of $\eta$. Now observe that $\tT^{-1}(B)_y = \{ u : (y, u) \in \tT^{-1}(B) \}
  = \G(y, \cdot)^{-1} ( B_y)$. We can continue the above calculation as follows:
  \begin{equation*}
    \begin{aligned}
      \int_B \nu( \dd z)
      = \int_\Y \eta_\tU( \tT^{-1}(B)_y) \nu_\Y (\dd y)
      = \int_\Y \eta_\tU\big( \G(y, \cdot)^{-1}(B_y) \big) \nu_\Y (\dd y)
       = \int_\Y \G(y, \cdot)_\sharp \eta_\tU(B_y) \nu_\Y(\dd y).
  \end{aligned}
\end{equation*}
Thus we have established that $\G(y, \cdot)_\sharp \eta_\tU$ is a conditional measure for $\nu$.
The desired result now follows from \Cref{prop:product-conditional-measures}(b) stating
the essential uniqueness of conditional measures.
\end{proof}

We highlight the simplicity of the proof and the generality of the conditions in \Cref{thm:triangular-maps-condition}. We do not require any assumptions such as regularity or monotonicity of the map $\T$ beyond the parameterization \eqref{transport-map}, and we certainly do not require $\eta$ and $\nu$ to be defined on the same spaces. The main restriction of our result is the assumption that $\eta = \eta_\tY \otimes \eta_\tU$, but this is an assumption on the reference measure, which can be chosen with considerable freedom.

The existence of the maps $\F$ and $\G$ may appear non-trivial at first sight. However, upon inspection
  of the proof of \Cref{thm:triangular-maps-condition} we realize that the map $\F$ is to some extent
  innocuous. For example, we can simply choose $\tY = \Y$ and $\eta = \nu_\Y \otimes \eta_\tU$ and take
  $\F$ to be the identity map. Indeed, this is the approach taken in \cite{carlier2016vector, ray2022efficacy, zhou2022deep}.
  If the above choice for $\eta$ is infeasible we can still take $\F$ to be any map that transports
  $\eta_\tY$ to $\nu_\Y$. An obvious choice would be a Brenier optimal transport map, which exists
  under very general assumptions on $\eta_\tY$ \cite{villani-OT}. Thus we focus our attention on the map $\G$ for
  the remainder of this subsection. First, we demonstrate that $\G$ can be identified in closed form in certain settings.

  \begin{example}[Gaussian random variables] \label{example:gaussian_case}
Let $\eta$ and $\nu$ be multivariate Gaussian measures where the conditional distribution $\nu(\cdot|y)$ has mean $m_{y}$ and covariance $\Sigma_{y}$ and $\eta_\tU$ is standard Gaussian. Let $\G$ be the affine transport map $\G(y,v) = m_{y} + \Sigma_{y}^{1/2}v$, where $\Sigma_{y}^{1/2}$ is any matrix square root. Then, $v \mapsto \G(y,v)$ pushes forward $\eta_\tU$ to the conditional measure $\nu(\cdot|y)$. 
\end{example}

  \begin{example}[Invertible transformations of random variables] \label{example:invertible_transformations}
    Consider measures of the form
     $ \nu = \Law \{ (y, u) | y = h(u - \xi), \quad u \sim \pi_1, \ \xi \sim \pi_2\},$
    where $\pi_1, \pi_2  \in \PP(\U)$. If $h^{-1}$ exists, then we can readily verify that $u = h^{-1}(y) + \xi$. In other
    words, $\nu(u | y) = \pi_2(u - h^{-1}(y))$. Now take $\eta = \nu_\Y \otimes \pi_2$ and observe that
    $\G(y, v) = v + h^{-1}(y)$ is the desired transport map.
\end{example}

A natural question arises regarding the existence of $\G$. While identification of $\G$ is  non-trivial, 
the existence of such a map  can be guaranteed under very general assumptions, following classic results from
probability theory. Below we consider the case where $\F = \text{Id}$.

\begin{proposition}\label{prop:existence-of-G}
  Take $\eta = \nu_\Y \otimes U[0,1]$ where $U[0,1]$ denotes the uniform measure on the interval $[0,1]$.
  Then there exists a measurable map $\G: \Y \times [0,1] \to \U$ so that $\G(y, \cdot)_\sharp U[0,1] = \nu(\cdot | y)$
  for any target $\nu \in \PP(\Z)$.
\end{proposition}

\begin{proof}
  Recall that we equipped $\Z$ with the Borel $\sigma$-algebra and that the system of conditional measures
  $\nu(\cdot | y)$ are by definition transition kernels. Then a direct application of \cite[Lem.~2.22]{kallenberg}
  gives the desired result.
\end{proof}

We can extend the above result by replacing the uniform measure $U[0,1]$ with another Radon
measure $\eta_\tU \in \PP(\tU)$.\footnote{Indeed \cite{zhou2022deep} presents a similar extension for the case of a Gaussian reference measure following the same line of thinking.} This uses the fact that
Borel measures on Polish spaces  are isomorphic to Borel measures on $[0,1]$; see \cite[Ch.~1 and Theorem~A1.6]{kallenberg}. Thus, the existence of $\G$ is not
an issue in our separable Banach space setting. More delicate questions arise, however, such as the characterization and
regularity of the maps $\G$ (and subsequently $\T$), which are important for approximation. We address some 
of these questions in the next subsection.

\subsection{Variational characterization of block triangular maps}
\label{subsec:existence-of-maps}

We now turn our attention to identifying  block triangular maps $\T$ of the form
\eqref{transport-map} via variational formulations, with a view towards practical
algorithms. Any computational approach will require approximation of the map $\T$ with a sequence of maps
$\T^n$ and so we need a notion of convergence for the resulting pushforwards to
the conditional measures $\nu( \cdot| y)$. 
We obtain such a result below under the assumption that the target $\nu$ is non-degenerate.
Recall the following definition:
\begin{definition}
  A measure $\mu \in \PP(\Z)$ is non-degenerate if for any collection of bounded linear functionals $\ell_1, \dots, \ell_n \in \Z^\ast$ (the dual of $\Z$) the measures $(\ell_1, \dots, \ell_n)_\sharp \mu \in \PP(\mathbb{R}^n)$
  are absolutely continuous. 
\end{definition} 

We now obtain the following convergence result in the setting where  the approximating sequence $\T^n_\sharp \eta$ converges to $\nu$ weakly 
on the product space. 
\begin{theorem}\label{thm:averaged-weak-convergence}
  Consider a reference measure $\eta = \eta_\tY \otimes \eta_\tU  \in \PP(\tZ)$ and
  a non-degenerate target measure
  $\nu \in \PP(\Z)$.
   Let $\{\T^n\}_{n \ge 0}$ be a sequence of maps of the form  \eqref{transport-map}
   with component maps $\F^n, \G^n$ as in \eqref{F-G-map-definition}.
   Furthermore, suppose that $\T^n_\sharp \eta \to \nu$ weakly as $n \to \infty$. Then, 
   for any $r > 0$, $y^\ast \in \Y$, and $f \in C_b(\U)$ it holds that
   \begin{equation*}
     \lim_{n \to \infty} \int_{B_r(y^\ast)} \int_\U f(u) \G^n(y, \cdot)_\sharp \eta_\tU(\dd u) \F^n_\sharp\eta_\tY(\dd y)
     \to \int_{B_r(y^\ast)} \int_\U f(u) \nu(\dd y, \dd u),
   \end{equation*}
   where $C_b(\U)$ denotes the space of continuous and bounded functions on $\U$.
\end{theorem}

\begin{proof}
 By definition of weak convergence we have for $(g,f) \in C_b(\Y) \times C_b(\X)$ that
  \begin{equation*}
    \int_\Y g(y)  \int_\U f(u) \G^n(y, \cdot)_\sharp \eta_\tU(\dd u) \F^n_\sharp \eta_\tY(\dd y)
    \to \int_\Y g(y) \int_\U f(u) \nu(\dd y, \dd u).
  \end{equation*}
  It further follows from \cite[Lem.~6.1]{agapiou2018sparsity} that since
  $\nu$ is non-degenerate then 
  open and convex sets are continuity sets of $\nu$. The desired result then follows
  from an application of the Portmanteau theorem \cite[Cor.~8.2.10]{bogachev2}.
{}
\end{proof}

We view  the above theorem as an averaged weak convergence for the conditionals, i.e.,
integrals of bounded and continuous functions with respect to $\G^n(y^\ast, \cdot)_\sharp \eta_\tU$
converge to integrals with respect to $\nu( \cdot | y^\ast)$ so long as we average those
integrals over small balls around $y^\ast$. One can obtain stronger convergence results
such as $y$-a.s.\thinspace convergence of $\G^n(y, \cdot)_\sharp \eta_\tU$ to $\nu(\cdot |y)$
by imposing stronger conditions on $\nu$ using general convergence results
for conditional measures; see  \cite{crimaldi2005convergence, goggin1994convergence}. We
do not pursue this direction at the moment since the required conditions on $\nu$ are difficult to verify in practice.

  We also note that the assumption of non-degeneracy on the target $\nu$ can
  be replaced with other (possibly more relaxed) conditions.
  For example, we only need $B_r(y^\ast)$  to be a continuity set of $\nu_\Y$.
  Alternatively, if $\nu$ is degenerate we can always relax the statement of
  \Cref{thm:averaged-weak-convergence} to a convergence result of the form
    \begin{equation*}
     \lim_{n \to \infty} \int_{\Y} g(y) \int_\U f(u) \G^n(y, \cdot)_\sharp \eta_\tU(\dd u) \F^n_\sharp\eta_\tY(\dd y)
     \to \int_{\Y} g(y) \int_\U f(u) \nu(\dd y, \dd u),
   \end{equation*}
   where $g$ is a Lipschitz approximation to the indicator of $B_r(y^\ast)$---for example,
   \begin{equation*}
     g(y) \coloneqq  \left\{
       \begin{aligned}
         &1 && y \in B_r(y^\ast), \\
         &\max\left\{ 0,  1 - \frac{\| y - y^\ast \|_\Y - r}{\eps} \right\} && y \in B_r(y^\ast)^c,
       \end{aligned}
       \right.
     \end{equation*}
     for a small parameter $\eps >0$. 

The above approximation result motivates a variational characterization of the block triangular maps $\T$ (along with their approximations $\T^n$) by minimizing statistical divergences.
If the chosen divergence metrizes weak convergence, one can then directly apply
\Cref{thm:triangular-maps-condition} to obtain convergence of expected values of
quantities of interest. This line of thinking leads us to optimization
 problems of the form \eqref{opt-prob-intro}. We
 recall the definition of a statistical divergence. 
\begin{definition}\label{def:statistical-divergence}
  A function $\D\colon \PP(\Z) \times  \PP(\Z) \to \mathbb{R}$
  is called a {\it statistical divergence} (or simply a divergence)
  on $\PP(\Z)$
  if for $\mu_1, \mu_2 \in \PP(\Z)$ it holds that 
  \begin{enumerate}
  \item $\D(\mu_1, \mu_2) \ge 0$.
    \item $\D(\mu_1, \mu_2) =0$ if and only if $\mu_1 = \mu_2$. 
  \end{enumerate}
\end{definition}
We now pose the following optimization problem,
\begin{equation}
  \label{abstract-optimization-problem}
  \minimize_{\T \in \mT} \D(\T_\sharp \eta, \nu),
\end{equation}
where $\mT$ is the space of measurable maps $\T$  parameterized as in \eqref{transport-map} and
\eqref{F-G-map-definition}. That is, 
\begin{equation}\label{triangular-map-space}
  \mT \coloneqq  \left\{ \T: \tZ \to \Z \: : \: \T(\ty,\tu) = \big(\F(\ty), \G(\F(\ty), \tu) \big) \ \text{for} \ 
  \F: \tY \to \Y, \ \G: \Y \times \tU \to \U \right \} 
\end{equation} 

\begin{remark}
It follows from \Cref{prop:existence-of-G} and the subsequent discussion that problem
\eqref{abstract-optimization-problem} has a global minimizer $\T^\dagger$ achieving
$D(\T^\dagger_\sharp \eta, \nu) = 0$ so long as we take $\eta = \eta_\tY \otimes \eta_\tU$. These minimizers, however, 
are not unique. For example, if we take $\tZ = \Z$ to be finite-dimensional Euclidean spaces
with an atomless reference measure $\eta$, then
the KR rearrangement $\T_{\text{KR}}$ serves as a global minimizer of \eqref{abstract-optimization-problem}. At 
the same time, letting $P$ be a block-diagonal permutation matrix that reorders the $\U$ (similarly the $\Y$) coordinates of $\Z$,
we can also 
construct the KR rearrangement $\T_{\text{KR}}'$ between the measures $P_\sharp \eta$ and $P_\sharp \nu$, and then 
$P^{-1} \circ \T' \circ P$ will also be a minimizer of \eqref{abstract-optimization-problem}.
Another example is the conditional Brenier maps of \Cref{prop:opt_maps}, which are fully block triangular 
as opposed to the KR maps that are strictly triangular.
\end{remark}

\subsection{Monotone block triangular maps}
\label{subsec:monotonicity-OT}

We now consider restricting the set $\mT$ in Problem~\eqref{abstract-optimization-problem} in a way that leads to unique minimizers, which also have the desirable regularity properties of OT maps.
To this end, we restrict our attention to the setting where $\tY = \Y$, $\tU = \U$, and hence $\tZ = \Z = \Y \times \U$, i.e., the reference measure $\eta$ and the target measure $\nu$ are defined on the same space. We also let $\Y$ and $\U$ be finite-dimensional Euclidean spaces. Motivated by \cite{carlier2016vector}
and existing literature on approximations of the KR rearrangement \cite{marzouk2016sampling},
we consider two subsets of  $\mT$:
\begin{itemize}
\item The set $\mT^M \subset \mT$ of monotone maps,
  \begin{equation*}
    \mT^M \coloneqq  \left\{ \T \in \mT \: :  \:  \left  (\T(z) - \T(z') \right )^\top ( z- z') \ge 0,
      \ \  \forall z,z' \in \Z\right\};
\end{equation*}
\item The set  $\mT^B \subset \mT$ of maps for which $\F$ and $\G$ are gradients of convex functions,
    \begin{equation*}
      \begin{aligned}
        \mT^B \coloneqq  \big\{ \T \in \mT \: : \:
        &  \exists f: \Y \to \R,   \ \  \exists g: \Y \times \U \to \R \text{ such that }  \\
        & y \mapsto f(y) \text{ and } v \mapsto g(y,v) \text{ are convex and  } \\
        & \F(y) = \nabla_y f(y),   \ \  \G(y, v) = \nabla_v g(y, v)        \big\}.
      \end{aligned}
\end{equation*}
\end{itemize}

The spaces $\mT^M$ and $\mT^B$ are closely related but are not the same. Elements of $\mT^B$
are monotone, but $\mT^B \subset \mT^M$ due 
to  a well-known result of Rockafellar
stating that maximal cyclically monotone maps $\T$ are uniquely determined by  gradients of proper
convex functions \cite[Thm.~24.8, 24.9]{rockafellar2015convex}. Cyclic monotonicity is a stronger condition than monotonicity and so there are monotone maps that are not gradients of convex functions.

In \Cref{subsec:regularization} we develop regularization techniques using  the space
$\mT^M$, but we note that $\mT^B$ is more convenient for our theoretical analysis.
Using $\mT^M$ leads to a natural penalty term that is convenient in practice and can be implemented with minimal
restrictions on our parameterization of the maps, while $\mT^B$ motivates the use of parameterizations for convex functions~\cite{amos2017input}. In either case, the monotonicity of the minimizers leads to desirable 
uniqueness and regularity properties. In particular, the Browder--Minty theorem  tells us that
continuous, bounded and coercive monotone maps are surjective \cite{zeidler2013nonlinear}, and this surjectivity allows us
to overcome issues with overfitting, which is also referred to as \emph{mode collapse} in the generative modeling literature.
We now show a uniqueness result for (\ref{abstract-optimization-problem}) when the space $\mT^B$ is utilized. Let us start by recalling \cite[Thm.~2.3]{carlier2016vector}. 
We emphasize that we only consider the case where $\tU = \U$, and hence eliminate the notation 
for the reference space of the parameter. 

\begin{proposition} \label{prop:opt_maps}
  Let $\Y$ and $\U$ be finite-dimensional Euclidean spaces. Consider a target measure $\nu \in \PP(\Y \times \U)$ 
  and a reference measure $\eta \in \PP(\Y \times \U)$. Assume the following:
  \begin{enumerate}[(i)]
  \item The reference measure $\eta \in \PP(\Y \times \U)$ has the form
    $\eta = \nu_\Y \otimes \eta_{\U}$. 
    \item The reference marginal $\eta_{\U}$ has a Lebesgue density with convex support on $\U$. 
  \item For each $y \in \Y$, the target conditional measure $\nu( \cdot | y)$ admits a
    Lebesgue density.
  \item $\int_\Y \int_\U \| u\|^2 \nu( \dd y, \dd u)  < \infty$ and
    $\int_{\U} \|u \|^2 \eta_{{\U}}( \dd u) < \infty$. 
    \end{enumerate}
  Then there exists a \emph{unique} map $\G = \nabla_u g(y, u)$, where $u \mapsto g(y,u)$ is convex $\forall y \in \Y$, such that 
  \begin{align} \label{eq:conditionalpushforwards}
      \G(y, \cdot)_\sharp \eta_{\U} = \nu(\cdot | y), \  \ \text{for } \nu_{\Y}\text{-}a.e.\ y.
  \end{align} 
  Moreover, this $\G$ minimizes the quadratic cost $\mathcal{M}(\S) \coloneqq \mathbb{E}_{(y,u) \sim \eta}\|u - \S(y,u)\|^2$ among all maps $\S: \Y \times \U \to \U$ that satisfy \eqref{eq:conditionalpushforwards}.
\end{proposition}

Following \cite{carlier2016vector}, we call the map identified by \Cref{prop:opt_maps} a ``conditional Brenier map.'' Next, combining this result with \Cref{thm:triangular-maps-condition}, we obtain a uniqueness result for optimization problems of the form \eqref{abstract-optimization-problem} over the set $\mT^B$.

\begin{theorem}\label{thm:brenier-uniqueness-map}
Let $\Y$ and $\U$ be finite-dimensional Euclidean spaces. Consider a target measure $\nu \in \PP(\Y \times \U)$, 
  and a reference measure $\eta \in \PP(\Y \times \U)$ of the form $\eta = \eta_\Y \otimes \eta_\U$.
  Suppose $\eta_\Y$ has no atoms and that 
  conditions (ii--iv) of
  \Cref{prop:opt_maps} are satisfied.
  Then, the optimization problem 
  \begin{equation*}
    \minimize_{\T \in \mT^B} \D(\T_\sharp \eta, \nu),
  \end{equation*}
  has a unique minimizer $\T^\dagger$
  achieving  $\D(\T^\dagger_\sharp \eta, \nu) = 0$ and  $\G^\dagger(y, \cdot)_\sharp \eta_\U = \nu( \cdot | y)$ for $\nu_\Y$-a.e.\ $\y$. %
\end{theorem}

\begin{proof}
  We begin by showing the existence of a minimizer $\T^\dagger$ which achieves $\D(\T^\dagger_\sharp \eta, \nu) =0$.
 Since we assumed $\eta_\Y$ has no atoms it follows from  the celebrated result of McCann \cite[Main Thm.]{mccann1995existence} that  there exists
  a unique map $\F^\dagger\colon \Y \to \Y$ which is the gradient of a convex function
  and $\F^\dagger_\sharp \eta_\Y  = \nu_\Y$. Thus $(\F^\dagger \times \Id)_\sharp \eta = \nu_\Y \otimes \eta_\U$.
  Let $\G^\dagger$ denote the unique monotone map from \Cref{prop:opt_maps} and
  define $\tT\colon (y,u) \mapsto (y, \G^\dagger(y, u) )$.  Now observe that   
  $\T^\dagger = \tT \circ (\F^\dagger \times \Id)$ satisfies $\T^\dagger_\sharp \eta = \nu$ and belongs to $\mT^B$
  by construction.

  We now verify the uniqueness of $\T^\dagger$. Suppose there exists another map $\T' \in \mT^B$,
  with components $\F', \G'$ given by gradients of convex functions, and such that
  $\T'_\sharp \eta = \nu$. By definition the $\Y$ marginal of $\T'_\sharp \eta$ coincides with
  $\nu_\Y$ and so $\F'_\sharp \eta_\Y = \nu_\Y$ thanks to the product structure of $\eta$.
  It follows from the uniqueness of $\F^\dagger$ that we should have $\F' = \F^\dagger$.
  On the other hand, \Cref{thm:triangular-maps-condition} implies that
  $\G'(y, \cdot)_\sharp \eta_\U = \nu( \cdot| y)$ for $\nu_\Y$-a.e.\ $\y \in \Y$. It follows from \Cref{prop:opt_maps}
  that we should have $\G'=\G^\dagger$, which yields the desired result.  
\end{proof}

\begin{remark}
  We  observe in \Cref{sec:numerics_OT} that using the space $\mT^M$
  still produces numerical solutions that approach the OT map of \Cref{thm:brenier-uniqueness-map}, meaning that these minimizers over $\mT^M$ are ``close to'' gradients of convex functions. Characterizing this behavior is an interesting direction of future research.
\end{remark}

 \begin{remark}
   We highlight that \Cref{thm:brenier-uniqueness-map} applies only to the finite-dimensional
   setting and in particular when $\tZ = \Z$. Our numerical algorithms in \Cref{sec:MGAN}
   and some of the experiments in \Cref{sec:numerics} will extend outside of these settings. Thus, there are still
   gaps in our theory that pose interesting directions for future research. An extension of \Cref{thm:brenier-uniqueness-map}
   to the infinite-dimensional setting will require extending \Cref{prop:opt_maps},
   which in turn relies heavily on McCann's result. Existence and uniqueness of Brenier maps
   in infinite dimensions is a contemporary topic in OT 
   and is only known under certain assumptions on the underlying spaces
   and on the reference and target measures \cite{ambrosio2005gradient, feyel2004monge, kolesnikov2014continuity}. 
 \end{remark}

 \section{The monotone GAN framework}\label{sec:MGAN}

This section develops 
 a practical framework for solving problems of the
form \eqref{opt-prob-intro}, motivated by the analysis
of \Cref{sec:theory}. We primarily focus on settings
where we only have access to samples from the reference $\eta$ and the target  $\nu$.
Henceforth, we write $\widehat{\eta}^N$, $\widehat{\nu}^N$ to denote empirical approximations to the
respective measures with $N$ i.i.d.\ samples. That is,
\begin{equation*}
  \begin{aligned}
    \widehat{\eta}^N &\coloneqq  \frac{1}{N} \sum_{j=1}^N \delta_{\tz_j}, \qquad
    \widehat{\nu}^N & \coloneqq  \frac{1}{N} \sum_{j=1}^N \delta_{z_j}, \qquad  \tz_j \iidsim \eta, \qquad z_j \iidsim \nu. 
\end{aligned}
\end{equation*}
We note that the methodology presented in this section will readily generalize to settings
where different number of samples are available from $\eta, \nu$, but we choose
to take $N$ samples from both measures for simplicity of presentation. We propose
 to approximate  \eqref{opt-prob-intro} 
\begin{equation}
  \label{practical-approximation-problem}
  \min_{\T \in \mT_\theta} \max_{g \in \mF_\omega}  \J(\T_\sharp \widehat{\eta}^N, \widehat{\nu}^N; g) +
   \R(\T; \widehat{\eta}^N),
\end{equation}
where $\mT_\theta \subset \mT$ is a space of block triangular maps of the form \eqref{triangular-map-space},
parameterized by $\theta$; $\mF_\omega$ is an appropriate space of functions
$g\colon \Z \to \R$ known as \emph{discriminators}, parameterized by $\omega$;
and $\R\colon \mT_\theta \times \PP(\Z) \to \mathbb{R}$ is a regularization functional.
The functional $\J\colon \PP(\Z) \times \PP(\Z) \times \mF_\omega \to \mathbb{R}$  
is chosen so that $ \max_{g \in \mF_\omega} \J(\cdot \,, \cdot \, ;  \, g)$ approximates a distance measure $\D$ such as an integral probability
metric or $f$-divergence~\cite{muller1997integral, birrell2022f}.
Our interest in such divergences stems from their successful deployment in 
large scale problems in ML, particularly in vision \cite{karras2019style}.
We will further discuss the choice of $\J$ and the role of the discriminator in \Cref{subsec:choosing-divergences}.

Let $\T_\theta^\dagger(w,v) = \big( \F_\theta^\dagger(w), \G_\theta^\dagger( \F_\theta^\dagger(w), v) \big)$ denote a minimizer of \eqref{practical-approximation-problem}, suppressing its dependence on $N, \mF_\omega$ and $\R$. Our hope is that  
$\G^\dagger_\theta$ is a good approximation to a true conditioning map $\G^\dagger$, such as the map from \Cref{thm:brenier-uniqueness-map}. We can then approximately sample the conditional measure $\nu( \cdot | y^\ast)$ for a fixed $y^\ast \in \Y$ by 
 drawing $v_j \iidsim \eta_\tU$ and 
 evaluating $u_j = \G_\theta^\dagger(y^\ast, v_j)$.

 In the remainder of this section we outline the details of our conditional simulation framework based on solving
 \eqref{practical-approximation-problem}. We discuss neural network parameterizations of $\T$ and $g$ in \Cref{subsec:NN-discretization},
 followed by our choices of regularization $\R$ in \Cref{subsec:regularization} and of the objective functional $\J$ in \Cref{subsec:choosing-divergences}. We summarize our algorithm in \Cref{subsec:algorithms},
 followed by a discussion of how our approach can be used to solve likelihood-free Bayesian inference problems in \Cref{subsec:char-BIP}.

\subsection{Neural network parameterizations of $\mT_\theta$
  and $\mF_\omega$}\label{subsec:NN-discretization}
Let $\X, \mH$ be  separable Banach  spaces. Below we define the notion of a neural network
mapping $\X \to \mH$; our construction is inspired by 
general families of neural operators such as those found in \cite{lifourier, lu2021learning, kovachki2023neural}. 
Let $\sigma\colon \RR \to \RR$ be a fixed function, henceforth referred to as an {\it activation
  function}. We overload our notation by writing
$\sigma(\bv) = ( \sigma(v_1), \dots \sigma(v_k) )^\top \in \RR^k$ for any vector $\bv \in \RR^k$. Let us fix an integer $L \ge 1$ (i.e., the depth parameter), the vector of integers
$\bd = (d_0, d_1, \dots, d_L) \in \NN^{L}$ (i.e., the width parameters), activation functions $\sigma^{(\ell)}: \RR \to \RR$, 
 matrices $\bW^{(\ell)} \in \RR^{d_\ell \times d_{\ell-1}}$
(i.e., the weights), vectors $\bb^{(\ell)} \in \RR^{d_\ell}$ (i.e., the biases), and bounded
linear operators $\Psi^{I}\colon \X \to \RR^{d_0}$ and $\Psi^{O}\colon \RR^{d_L} \to \mH$.
We then say that a map $\Q_\alpha\colon \X \to \mH$ is a
neural network if it has the form
\begin{equation*}
  \Q_\alpha(x) =  \Psi^{O}  \bigg( \sigma^{(L)} \left( \bW^{(L)} \bx^{(L)} + \bb^{(L)} \right) \bigg), \quad
  \bx^{(\ell)}= \sigma^{(\ell)}\left( \bW^{(\ell)} \bx^{(\ell-1)} + \bb^{(\ell)} \right), \quad
  \bx^{(0)} = \Psi^{I}(x),
\end{equation*}
where we use $\alpha \coloneqq  \{ (\bW^{(\ell)}, \bb^{(\ell)}) \}_{\ell=1}^L$ to denote
the collection of weights and biases of the neural network $\Q_\alpha$. Furthermore,
we refer to the collection of integers $L, d_0, \dots, d_L$ together with 
activation functions $\sigma^{(1)},\dots, \sigma^{(L)}$ and
the operators $\Psi^I, \Psi^O$
as the {\it architecture}
of $\Q_\alpha$. To this end, we define the spaces of neural networks
sharing the same architecture as
\begin{equation*}
  \mQ_\alpha(\X, \mH; \A) \coloneqq  \left\{ \Q_\alpha: \X \to \mH \: \big| \: \Q_\alpha \  \text{has architecture} \ 
  \A=\{ L, \bd, \sigma^{(1)}, \dots, \sigma^{(L)}, \Psi^I, \Psi^O\} \right\}.
\end{equation*}
With the above notation, we then consider the spaces  for the maps $\T$ and
the discriminator $f$ given by
\begin{equation*}
  \begin{aligned}
    \mF_\omega & = \mQ_\omega( \Z; \R; \A_0), &  && \\
    \mT_\theta &= \Big\{ \T: \tZ \to \Z, \: \theta =(\theta_1, \theta_2)  \   \big |  && \: \T(w,v) = \big ( \F_{\theta_1}(w), \G_{\theta_2}(\F_{\theta_1}(w), v)  \big ), \\
               & & &  \  \F_{\theta_1} \in \mQ_{\theta_1}(\tY; \Y; \A_1) , \G_{\theta_2} \in \mQ_{\theta_2} (\Z; \U; \A_2) \Big\}.
 \end{aligned}
\end{equation*}
For brevity, the architectures $\A_0, \A_1, \A_2$ are suppressed in our notation for $\mF_\omega, \mT_\theta, \mT_\theta^B$, and will be identified on a case-by-case basis for the  numerical experiments in \Cref{sec:numerics}.

 \begin{remark}
   We note that our choice of the space $\mT_\theta$ is completely generic and does not impose any form of monotonicity on the components $\F_{\theta_1}, \G_{\theta_2}$. We make this choice to have maximal flexibility in the design of architectures
   and to allow practitioners to utilize existing optimal architectures for the task at hand. Instead, we impose monotonicity using a penalty term that is outlined in \Cref{subsec:regularization}. An alternative
   approach would be to directly parameterize $\F_{\theta_1}$, $\G_{\theta_2}$ as gradients of (partially)
   convex functions \cite{amos2017input}. Indeed, such parameterizations have already been used
   for approximation of OT maps \cite{huang2020convex, makkuva2020optimal, onken2021ot}, but the expressivity of
   the associated input-convex neural networks and the design of appropriate architectures  in high-dimensional examples are not well understood, so we do not utilize these constructions here. 
   \label{remark:icnns}
 \end{remark}

 \subsection{Monotonicity penalty}\label{subsec:regularization}
In order to make our maps $\T_\theta$ (approximately) monotone 
 we propose to regularize~\eqref{practical-approximation-problem}
 using an average monotonicity penalty. More precisely, let $\Z = \Y \times \U$ as before and suppose that $\tZ = \Z$, i.e., that $\T$ maps $\Z \to \Z$.
 Then we propose the idealized penalty term
\begin{equation*} \label{eq:average_monotonicity}
  \R(\T; \mu) = - \lambda \EE_{z \sim \mu} \EE_{z' \sim \mu} \langle \T(z) - \T(z'), z - z' \rangle_{\Z},
\end{equation*}
for some positive constant $\lambda >0$ and any measure $\mu \in \PP(\Z)$. Including this regularization term in the minimization problem \eqref{practical-approximation-problem} 
in fact encourages $\T$ to be increasing (in the prescribed sense) over regions that $\mu$ endows with mass.
It is natural for us to take $\mu \equiv \eta$ so that
the regularization term encourages monotonicity of the map at inputs in the support of the reference distribution. In practice, one can set $\mu \equiv \widehat{\eta}^N$, or alternatively 
generate a new set of i.i.d.\ samples from the reference and use those samples to evaluate the penalty term.

It is important to note that while the average monotonicity penalty
does not ensure that $\T$ is monotone everywhere (not even on the support of $\mu$), numerically we find  
that this regularization term is sufficient to ensure that $\T$ is monotone with high probability.
In particular, one can easily compute an empirical approximation to  
\begin{equation}
\label{probability-monotonicity}
\mathbb{P}_{z \sim \mu, z' \sim \mu}
\big[ \langle \T(z) - \T(z'), z - z' \rangle_{\Z} >0 \big],
\end{equation}
that can be tracked during training as a proxy for the map's monotonicity over $\text{supp}(\mu)$.

 \subsection{Choosing the functional $\J$}\label{subsec:choosing-divergences}
The appropriate choice of the functional $\J$ is a compromise between
the computational cost of solving \eqref{practical-approximation-problem}
and the quality of the minimizer as an approximation to the solution of \eqref{abstract-optimization-problem}.
Various popular choices of the divergence $\D$  exist in the literature. Most
notably, the forward KL divergence (equivalently, maximum likelihood estimation)  
is widely used for finding 
invertible triangular maps and normalizing flows~\cite{papamakarios2019normalizing}, and the reverse KL divergence is used for 
variational inference or for training autoencoders~\cite{rezende2015variational, blei2017variational}. 
Other
possible choices include maximum mean discrepancy (MMD)~\cite{binkowski2018demystifying}, Wasserstein distances~\cite{arjovsky2017wasserstein}, and
$f$-divergences \cite{nowozin2016f, zhou2022deep}.

Our proposed framework is not tied to a specific choice of $\J$. Our theoretical
results suggest that so long as $\J$ yields a good approximation to a divergence, then
the resulting algorithm should be capable of approximating the map $\G^\dagger$ well. 
In our current setup, our main requirement on $\J$ is that it does not involve the pushforward density $\G(y,\cdot)_\sharp \eta_\U$, as computing this quantity would require evaluating the inverse of the map $u \mapsto \G(y,u)$ and its Jacobian determinant, which are computationally intensive.

Our choice of $\J$ in \eqref{practical-approximation-problem} can be 
motivated by the family of  \(f\)-divergences \cite{birrell2022f}:
Let $f\colon \Omega_f \subseteq \RR \to \RR$ be a continuous, convex function with $f(1) = 0$ and let $f^\ast\colon \Omega_{f^*} \subseteq \RR \to \RR$ denote its Legendre transform. 
We then define the $f$-divergence $\D_f\colon \PP( \Z) \times \PP(\Z) \to \RR$ as 
\begin{equation} \label{eq:IPM_fdiv}
    \D_f( \mu_1, \mu_2) = \sup_{g \in \mF} \int g(z) \mu_1(\dd z)  - \int f^\ast(g(z)) \mu_2(\dd z),
\end{equation}
where $\mF$ is a class of (measurable) functions mapping $\Z \to \Omega_{f^*}$, often referred to as 
the discriminator class in the terminology of generative adversarial models. We note that the standard
definition of a \(f\)-divergence, as in \cite{ali1966ageneral}, 
matches our definition in~\eqref{eq:IPM_fdiv} under the assumption $\mu_1$ and $\mu_2$ are equivalent measures, i.e., \(\mu_1 \sim \mu_2\)\footnote{This follows by a calculation similar 
to~\cite[Equation (4)]{nowozin2016f} with the observation that the continuity assumption on \(f\) allows
us to use \cite[Theorem 14.60]{rockafellar2005variational} to obtain equality, instead of a lower bound, between a integral probability metric and a $f$-divergence.}.
For specific choices of Legendre transforms $f^\ast$, we also observe that 
the objective functional $\max_{g \in \mF_\omega} \J(\T_\sharp \widehat{\eta}^N, \widehat{\nu}^N;g)$ 
in~\eqref{practical-approximation-problem} can be viewed as an approximation to the divergence 
$\D_f(\T_\sharp \widehat{\eta}^N, \widehat{\nu}^N)$, where the space $\mF$ is replaced by a (possibly parametric) 
class $\mF_\omega$.
We now outline some choices of $\D_f$ and the associated functional $\J$ that we 
used in our experiments in~\Cref{sec:numerics}.

\begin{itemize}
    \item Following \cite[Table~2]{nowozin2016f}, we can choose $f^\ast(t) = - \log( 2 - \exp(t))$, which leads to a %
    generalization of the Jensen-Shannon divergence $\D_f$. Moreover, reparameterizing 
the discriminator as $g(z) = \log 2 v(z) $ for some measurable function\footnote{Here, we mean measurable as being with respect to a measure to which both $\mu_1, \mu_2$ are absolutely continuous, e.g., $1/2 \mu_1 + 1/2 \mu_2$.} $v: \Z \to (0,1)$  yields 
\begin{equation*}
\begin{aligned}
    \D_{\text{GAN}}(\mu_1, \mu_2) & \coloneqq \sup_{v \in \mF} \int \log v(z) \mu_1(\dd z) 
    + \int \log  ( 1 - v(z) ) \mu_2 (\dd z) \\
    & \eqqcolon \sup_{v \in \mF} \J_{\text{GAN}}( \mu_1, \mu_2; v),
\end{aligned}
\end{equation*}
which is precisely the original GAN loss of~\cite{goodfellowGANS}, up to the additive constant $\log(4)$.

\item We may take $f^\ast (t) = t$ and $\mF$ to be the class of Lipschitz-1 functions on 
$\Z$ to obtain the dual formulation of the Wasserstein-1 distance
\begin{equation*}
    \D_{\text{W}_1}(\mu_1, \mu_2) \coloneqq \sup_{g \in \text{Lip}_1} 
    \int g(z) \mu_1(\dd z)  - \int g(z) \mu_2(\dd z).
\end{equation*}
In the case where $\Z$ is a finite-dimensional Euclidean space  
the Lipschitz-1 constraint on $g$ can be further relaxed to a gradient penalty (GP) to obtain 
\begin{equation*}
\begin{aligned}
    \D_{\text{WGP}}(\mu_1, \mu_2) & \coloneqq \sup_{g \in C^1(\Z)} 
    \int g(z) \mu_1(\dd z)  - \int g(z) \mu_2(\dd z) + \gamma \int  (\| \nabla g(z)\| - 1 )^2 \mu^\ast(\dd z) \\
    & \eqqcolon \sup_{g \in C^1(\Z)} \J_{\text{WGP}} (\mu_1, \mu_2; g),
\end{aligned}
\end{equation*}
where  $\gamma >0$ is a penalty coefficient. The measure $\mu^\ast$ is somewhat arbitrary,
but the following particular choice yields the  Wasserstein-GAN GP loss of \cite{gulrajani2017improved}:
\begin{equation*}
  \mu^\ast \coloneqq  \Law \big\{ z \: \big| \: z= \alpha z_1 + (1- \alpha) z_2, \quad \alpha \sim U[0,1], \:
  z_1 \sim \mu_1, \: z_2 \sim \mu_2 \big\}.
\end{equation*}

 \item In some of our low-dimensional examples we shall also use 
 the Least Squares GAN (\LSGAN) functional of~\cite{mao2017least} due to its simplicity and 
 the fact that it showed good empirical performance. That is,
  \begin{equation*}
    \J_{\LSGAN}( \mu_1, \mu_2; g) \coloneqq  \frac{1}{2}
    \left[ \int (g(z) - a) ^2 - (g(z) - b)^2 \mu_1(\dd z) - \int ( g(z) - c)^2 \mu_2(\dd z) \right],
  \end{equation*}
  where $a,b,c$ are scalar constants, which we simply chose as $a = c = 1$ and $b = 0$. Note that, unlike other loss functions, the \LSGAN loss 
  cannot be written as an $f$-divergence. In practice, this functional is optimized using alternating steps that maximize the last two terms with respect to $g$, and minimize the first term alone with respect to $\mu_2$.
\end{itemize}

  We reiterate that our above choices of $\J$ are driven by empirical success with numerical experiments in \Cref{sec:numerics}. The question of the optimal choice of $\J$
  and more broadly the divergence $\D$, however, is a contemporary topic in generative modeling and is the subject of intense research \cite{birrell2022f, nowozin2016f, birrell2022f}. Our goal is not to make a
  statement on the best choice of these functionals, but rather to focus on the transport methodology 
  for conditional simulation.

\subsection{Summary of the algorithm}\label{subsec:algorithms} 

We now present a summary of  the \MGAN training procedure, leading to \Cref{alg:main}, which is used in the numerical experiments of \Cref{sec:numerics}.
To learn the parameters $\theta, \omega$ for $\T,f$, %
we solve~\eqref{practical-approximation-problem} using an alternating gradient descent procedure that is common for training GANs~\cite{goodfellowGANS}. This approach repeats the following two steps: (1) Update the parameters $\theta$
while holding $\omega$ fixed; and (2) Update the parameters $\omega$ while holding $\theta$ fixed.
Informally, the first step improves the map $\T$ so that  the pushforward samples are closer to $\widehat{\nu}^N$.
The second step updates the discriminator $g$ to better distinguish
``real'' and ``fake'' samples from $\widehat{\nu}^N$ and $\T_\sharp\widehat{\eta}^N$, respectively. %
We note that when using the \WGANGP functional for the image in-painting example of \Cref{sec:image-painting}, \(\theta\) is updated
 once for every five updates of \(\omega\), as is standard practice for large-scale problems.

For conditional sampling, we are primarily interested in approximating the component function $\G$ of the map $\T$, which pushes forward $\eta_{\tU}$ to $\nu(\cdot|y)$.
Thus, for the experiments below, we choose $\F = \Id$, which implies that $\eta_{\tY} = \nu_{\Y}$. We also choose $\tU = \U$, but with $\eta_\tU \equiv \eta_\U \neq \nu_\U$ in general. To simplify notation, we thus write the product reference measure as $\eta = \nu_{\Y} \otimes \eta_{\U}$ as was done in \Cref{subsec:monotonicity-OT}.

At each gradient descent step, we replace the expectations in $\J$ with empirical averages over mini-batches from the reference measure $\widehat{\eta}^N$ and the target measure $\widehat{\nu}^N$, as in the standard GAN training procedure~\cite{goodfellowGANS}.
In particular, for each update of the parameters $\theta, \omega$, two mini-batches of size $M$ are sampled: %
one from the reference marginal $\eta_{\U}$, %
and one from the training set of the joint target $\widehat{\nu}^N$. We then form the two joint empirical measures $\widehat{\eta}^M$ and $\widehat{\nu}^M$ using the same samples $y_i \sim  \widehat{\nu}_\Y^N$. For some objective functionals (e.g., \LSGAN), we found good empirical performance by also sampling an independent mini-batch of size $M$ from the marginal $\widetilde{y}_i \sim \widehat{\nu}_\Y^N$ %
and forming the empirical measure $\widehat{\eta}^M$ from $\{(\widetilde{y}_i,v_i)\}_{i=1}^M$ where $v_i \sim \eta_{\U}$,
although this extra sampling step is not strictly necessary for unbiased estimates of the objective.
We update the parameters for multiple epochs (i.e., passes through the training set), until the evaluations of the functional $\J$ and the penalty $\R$ converge.

\begin{algorithm}[!htb]
\begin{algorithmic}[1]
  \STATE {\bfseries Input}: target samples $\{ (\y_j, \u_j) \}_{j=1}^N \iidsim \nu$,
   monotonicity penalty parameter \(\lambda > 0\), number of epochs, batch size $M$
  \STATE {\bfseries Output}: Mapping $\G \in \mQ_{\theta_2}$ satisfying $\G(y, \cdot)_\sharp \eta_\U \approx \nu(\cdot | \y)$ for any $\y \in \Y$

  \FOR{\text{number of epochs}}
  
  \STATE Sample mini-batch of size \(M\) from training set \((y_1, u_1),\dots,(y_{M}, u_{M}) \iidsim \widehat{\nu}^N\)\; 
  \STATE Draw \(M\) reference samples $v_1, \dots, v_M \iidsim \eta_\U$
  \STATE Form empirical measures $\widehat{\eta}^M$ and $\widehat{\nu}^M$
  from  $\{ (y_i, v_i) \}_{i=1}^M$  and $\{ (y_i, u_i) \}_{i=1}^M$ respectively
  \STATE Update $\theta$ in $\G$ by descending its stochastic gradient
   $\nabla_{\theta} \J(\T_{\sharp}\widehat{\eta}^M,\widehat{\nu}^M;g) + \R(\T;\widehat{\eta}^M)$\;
  \STATE Update $\omega$ in $g$ by ascending its stochastic gradient
   $\nabla_{\omega} \J(\T_{\sharp}\widehat{\eta}^M,\widehat{\nu}^M;g)$\;
  \ENDFOR
  \end{algorithmic}
  \caption{Outline of the \MGAN training procedure. \label{alg:main}}
\end{algorithm}

\subsection{Likelihood-free Bayesian inference}\label{subsec:char-BIP}

Since conditional simulation is a fundamental task in Bayesian inference, 
we use this section to illustrate how \MGANs can be used for likelihood-free Bayesian inference. To this end, let $\Y$ and $\U$ denote the data space and parameter space of interest, respectively, and let the conditional measure $\nu(\cdot | u)$ represent a statistical model for the data $y \in \Y$, parameterized by $u \in \U$. Furthermore, let $\nu_0 \in \PP(\U)$ denote a \emph{prior} measure on $u$. 
Then, the goal of Bayesian inference is to characterize the conditional measures $\nu(\cdot | y)$, which is the system of conditionals of the joint measure
$\nu(\dd y, \dd u) = \nu(\dd y | u) \nu_0(\dd u) \in \PP(\Y \times \U)$, where $\nu(\cdot | u)$ are regarded as transition kernels.

Let us now consider 
a reference measure $\eta  = \eta_\Y \otimes \eta_\U \in \PP(\Y \times \U)$.
Then from any $\T^\dagger$ that is a global minimizer of \eqref{abstract-optimization-problem}
we can extract $\G^\dagger$ and it follows from \Cref{thm:triangular-maps-condition} that
\begin{equation}\label{BIP-posterior-map-char}
  \G^\dagger(y, \cdot)_\sharp \eta_\U = \nu( \cdot | y) \qquad \text{for } \nu_\Y\text{-a.e. }y \in \Y. 
\end{equation}
In other words,  $\G^\dagger$ completely characterizes the posterior measure for all
values of  $y \sim \nu_\Y$. %

We make two key observations about the map. First, the identity \eqref{BIP-posterior-map-char} suggests that
in the case of Bayesian inverse problems it is reasonable to choose 
$\eta_\U = \nu_0$, i.e., to choose the prior as the reference measure on $\U$. This
choice is motivated by the fact that posterior measures often deviate from the prior (essentially) over a low-dimensional subspace~\cite{cui2014likelihood, marzouk-dili, bigoni2019greedy}. 
Hence, one expects that this choice of the reference would
result in a map $\G^\dagger$ that is also (essentially) low-dimensional and that captures how the posterior deviates from the prior. Second, similarly to other conditional generative models~\cite{papamakarios2016fast}, the map $\G^\dagger$ provides a single function for cheaply sampling from
the conditional $\nu(\cdot|\y)$ given any realization of the data $\y \in  \supp{\nu_\Y}$.
In comparison to traditional sampling
algorithms (e.g., MCMC) that must be repeated for each new realization of $\y$, the process of learning this single map
amortizes the cost of inference over the data.

To make our approach concrete, we simulate a set of samples $u_j \iidsim \nu_0$ 
and evaluate our forward model to obtain 
corresponding observations $y_j \iidsim \nu( \cdot | u_j)$. The tuples $(y_j, u_j)$ are then draws from the
target joint measure $\nu$. We draw an additional set of samples  $\tilde{u}_j \sim \nu_0$ 
so that $(y_j, \tilde{u}_j)$ are draws from the reference distribution $\eta =  \nu_{\Y} \otimes \nu_0$. 
We then apply the \MGAN approach to
identify the posterior using the aforementioned empirical samples. 
 
We remark that a likelihood function does not appear in the optimization problems
\eqref{abstract-optimization-problem} or \eqref{practical-approximation-problem}
and so we only need to evaluate the forward map when 
generating the samples $y_j \sim \nu(\cdot | u_j)$. %
In the case of PDE-constrained inverse problems (see for example \Cref{sec:darcy-flow}), simulating the measurements $y_j$ is the most costly
step in generating the data for \MGANs. Given that these samples are independent, however, they can be generated in parallel, which is
a major advantage over standard MCMC algorithms.

\section{Numerical experiments}\label{sec:numerics}

We now present a series of experiments that demonstrate the effectiveness of
\MGANs in various conditional sampling applications. %
In \Cref{sec:synthetic-example} we demonstrate the importance of the monotonicity constraint for accurate %
uncertainty quantification %
on nonlinear regression problems with non-Gaussian noise models. In \Cref{sec:banana} we show that block triangular maps are insensitive to variable ordering, in contrast with strictly triangular maps. %
\Cref{sec:numerics_OT} shows that the \MGAN framework can recover $L^2$ optimal transport maps. In \Cref{sec:ODE-inverse,sec:darcy-flow} we present applications to two inverse problems with non-Gaussian posterior measures: parameter inference in coupled ODEs and a Darcy flow model. Finally, \Cref{sec:image-painting} demonstrates the feasibility of \MGAN in high-dimensional conditional sampling problems arising in imaging. Unless otherwise stated, we take the reference measure to be $\eta = \nu_\Y \otimes N(0, I)$; that is, $\eta$ has the same $y$ marginal as the training data, allowing us to take $\F = \Id$, while the $u$ marginal is a standard Gaussian of the appropriate dimension.
Code to reproduce the numerical results is available online at: \url{www.github.com/baptistar/MGAN}.

\subsection{Synthetic examples}\label{sec:synthetic-example}

We start with a simple set of  synthetic examples  where the conditionals  $\nu(\cdot | y)$  can be computed  explicitly.
Consider the following input-to-output maps,
\begin{align}
u &= \tanh(y) + \xi, \qquad \xi \sim \Gamma(1,0.3), \label{eq:tanhmodel_out} \\
u &= \tanh(y + \xi), \qquad \xi \sim \mathcal{N}(0, 0.05), \label{eq:tanhmodel_in} \\
u &= \xi \tanh(y), \qquad \quad \: \xi \sim \Gamma(1,0.3), \label{eq:tanhmodel_mult} 
\end{align}
where \(y \sim U[-3,3]\) in all cases. 
We considered the problem of conditioning $u$ on $y$  and compared \MGAN  maps
computed via the \LSGAN loss functional using $N = 50,000$ training samples and with $\lambda = 0$ (i.e., no monotonicity penalty) and  $\lambda = 0.01$. %
We parameterized each map $\G$ as a three-layer, fully connected neural network
with hidden layer sizes \(256/512/128\) and leaky ReLU activation functions \cite{maas2013rectifier} with parameter $\alpha = 0.2$. We used 
the same architecture for our discriminator with an additional 
linear transformation in the final layer to make the output one dimensional.
Training was
performed using the Adam algorithm \cite{kingma2014adam} with learning rate $2 \times 10^{-4}$ and parameters $\beta_1 = 0.5$ and
$\beta_2 = 0.999$. We used a batch size of $M = 100$ and trained for $300$ epochs. 

\begin{figure}[!ht]
    \centering
    \hspace{.5cm}
    \begin{subfigure}[b]{0.23\textwidth}
      \begin{overpic}[width=\textwidth]{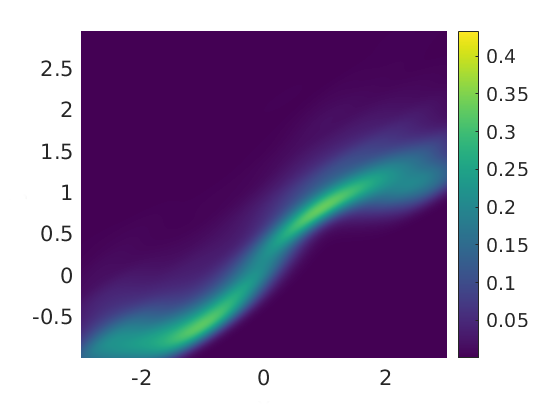}
        \put(-24, 37){\eqref{eq:tanhmodel_out}}
        \put(1, 39){{\tiny $u$}}
    \end{overpic}\\
    \begin{overpic}[width=\textwidth]{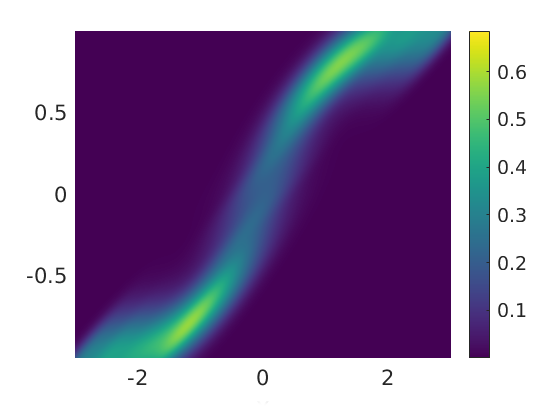}
       \put(-24, 37){\eqref{eq:tanhmodel_in}}
       \put(1, 39){{\tiny $u$}}
    \end{overpic}\\
    \begin{overpic}[width=\textwidth]{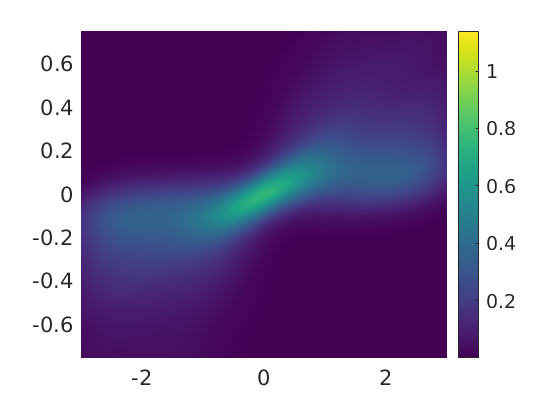}
       \put(-24, 37){\eqref{eq:tanhmodel_mult}}
       \put(1, 39){{\tiny $u$}}
        \put(45, 0.5){{\tiny $y$}}
    \end{overpic}
        \caption{$\nu$}
    \end{subfigure}
        \begin{subfigure}[b]{0.23\textwidth}
      \includegraphics[width=\textwidth]{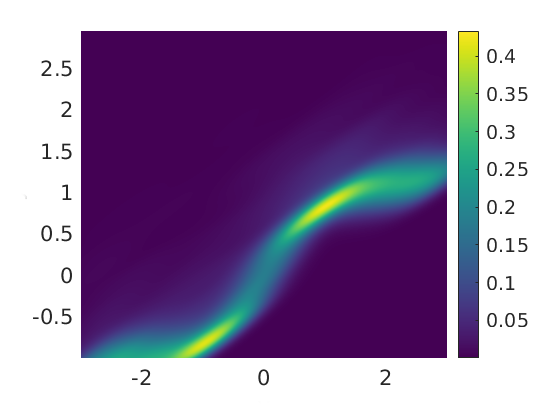}\\
      \includegraphics[width=\textwidth]{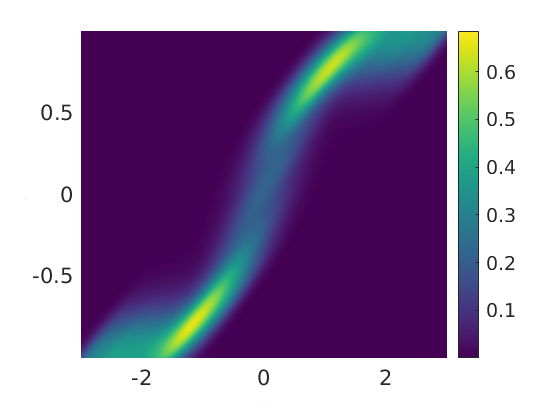}\\
        \begin{overpic}[width=\textwidth]{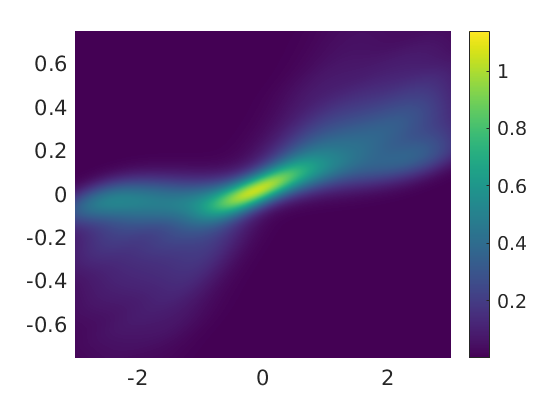}
        \put(45, 0.5){{\tiny $y$}}
        \end{overpic}
       \caption{$\lambda = 0$}
    \end{subfigure}
    \begin{subfigure}[b]{0.23\textwidth}
      \includegraphics[width=\textwidth]{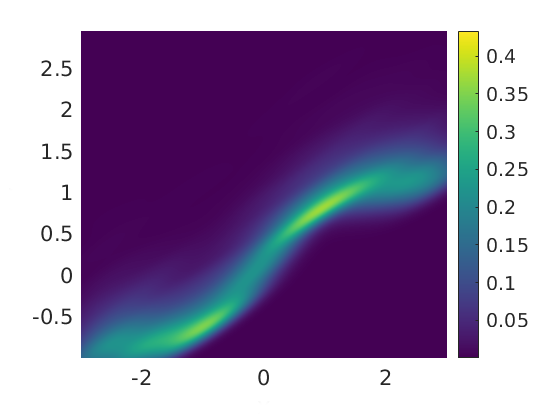}\\
      \includegraphics[width=\textwidth]{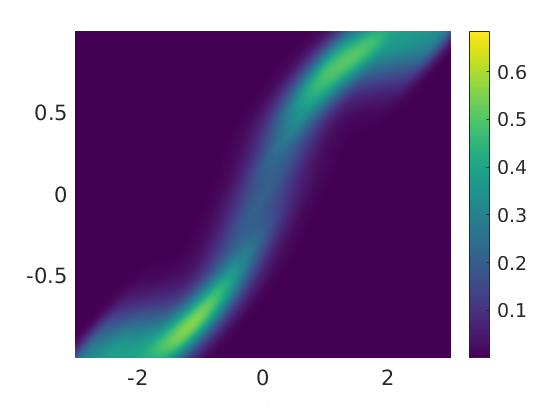} \\
      \begin{overpic}[width=\textwidth]{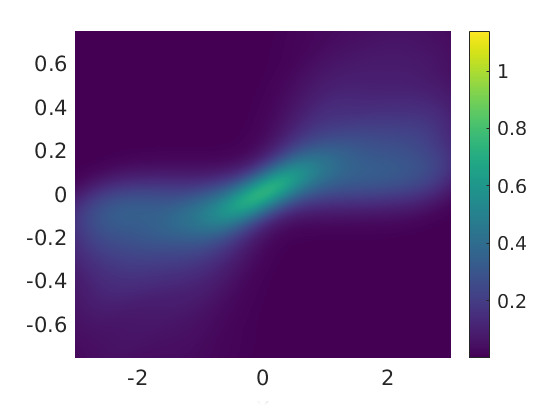}
        \put(45, 0.5){{\tiny $y$}}
        \end{overpic}
        \caption{$\lambda = 0.01$}
    \end{subfigure}
    \begin{subfigure}[b]{0.23\textwidth}
      \includegraphics[width=\textwidth]{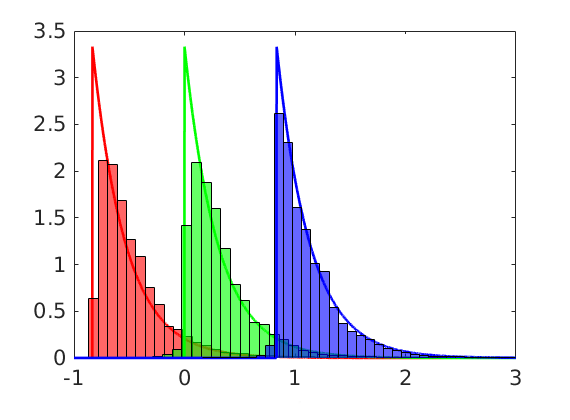}\\
      \includegraphics[width=\textwidth]{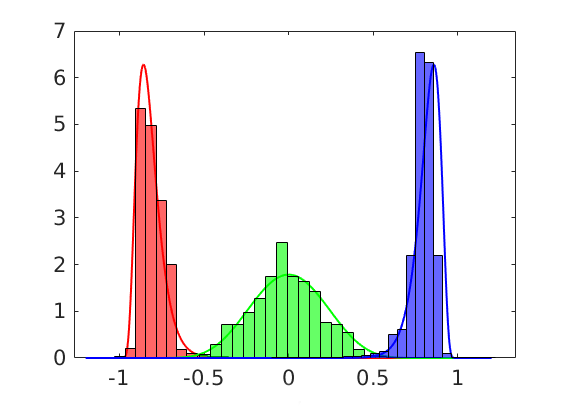}\\
     \begin{overpic}[width=\textwidth]{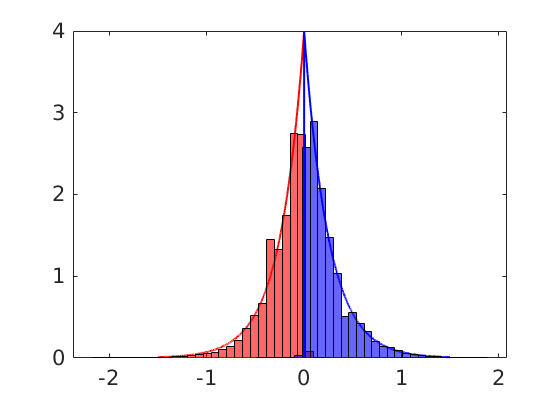}
        \put(52, 0.5){{\tiny $u$}}
        \end{overpic}
       \caption{$\nu(\cdot | y)$}
    \end{subfigure}
    \caption{
      The rows correspond to problems
      \eqref{eq:tanhmodel_out}, \eqref{eq:tanhmodel_in} and 
      \eqref{eq:tanhmodel_mult}, respectively. The first three columns
      compare the true joint densities for $\nu$ to kernel density
      estimates (KDEs) of conditional samples from 
      \MGAN with ($\lambda=0.01$) and without ($\lambda = 0$) the monotonicity penalty.   
      The last column compares histograms of conditional samples from
      \MGAN with $\lambda = 0.01$ to the true conditional densities (solid lines) for all three problems. The red, green and blue colors correspond to the distributions for $u | y = -1.1$, $u | y = 0$ and $u | y = 1.1$. 
}
    \label{fig:tanh_stats}
  \end{figure}

\Cref{fig:tanh_stats}(a--c) compares the true joint densities $\nu$ to the \MGAN
approximations with and without the monotonicity penalty. 
We observe a better match between the true density and the \MGAN pushforward $\T_\sharp \eta$ obtained with the monotonicity penalty, particularly in regions of high probability. 
\Cref{fig:tanh_stats}(d) compares histograms of conditional samples obtained from \MGAN to the
true conditional PDFs, explicitly showing \MGAN's ability to capture the conditionals correctly.

\begin{figure}[t]
    \centering
    \begin{subfigure}[b]{0.75\textwidth}
    \vspace{0.5cm}
      \begin{overpic}[width=\textwidth]{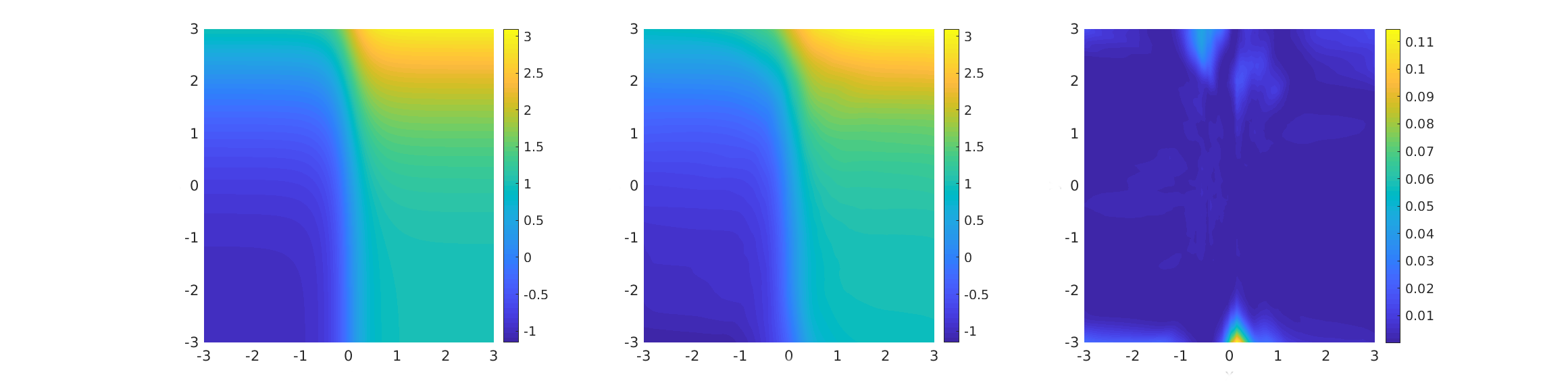}
        \put(0, 12){\eqref{eq:tanhmodel_out}}
        \put(20,25){\textrm{KR}}
        \put(44.5,25){\MdGAN}
        \put(74,25){Error}
        \put(9.5, 12.5){{\tiny $u$}}
      \end{overpic}
    \end{subfigure}

    \begin{subfigure}[b]{0.75\textwidth}
      \begin{overpic}[width=\textwidth]{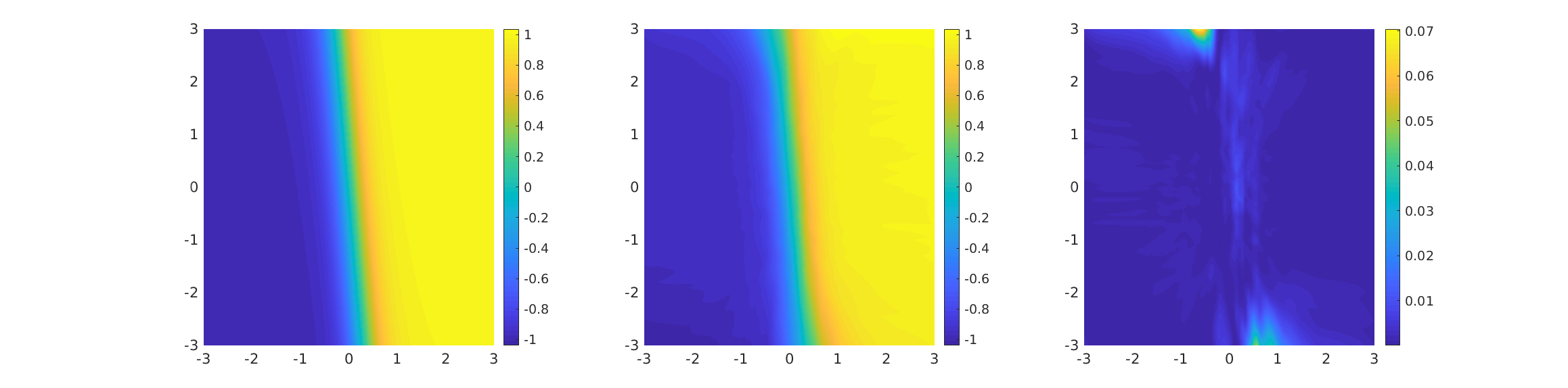}
        \put(0, 12){\eqref{eq:tanhmodel_in}}
        \put(9.5, 12.5){{\tiny $u$}}
      \end{overpic}
    \end{subfigure}

    \begin{subfigure}[b]{0.75\textwidth}
      \begin{overpic}[width=\textwidth]{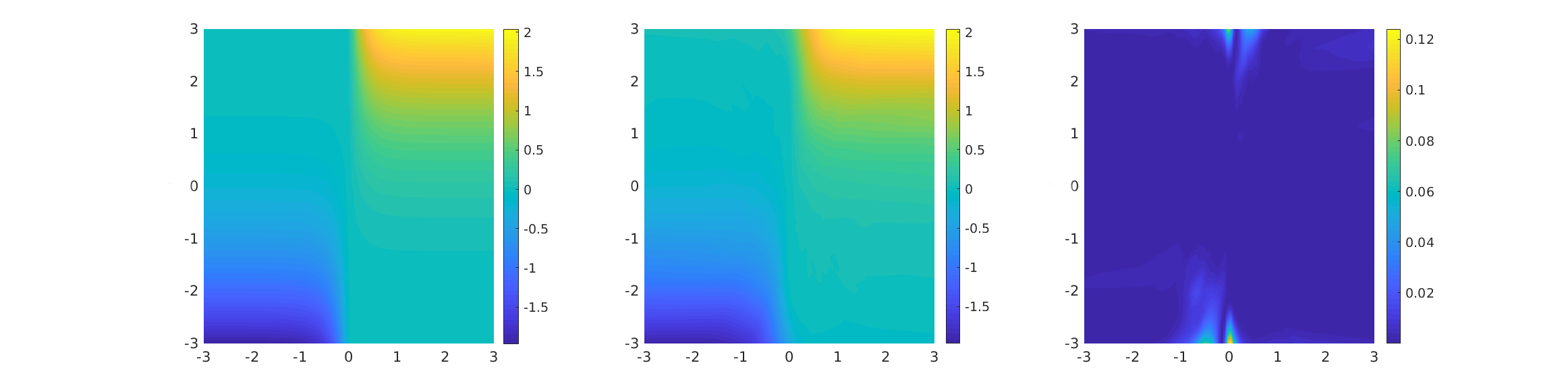}
        \put(0, 12){\eqref{eq:tanhmodel_mult}}
        \put(9.5, 12.5){{\tiny $u$}}
        \put(21.5, -0.5){{\tiny $y$}}
        \put(49.5, -0.5){{\tiny $y$}}
        \put(77.5, -0.5){{\tiny $y$}}
      \end{overpic}
    \end{subfigure}

    \caption{Each row corresponds to the problems \eqref{eq:tanhmodel_out}, \eqref{eq:tanhmodel_in} and \eqref{eq:tanhmodel_mult}, respectively.
      The first column shows the (second component function of) the true KR rearrangement, the second column shows the \MGAN map $\G$, and the last column shows the absolute pointwise
      error between the two.}
    \label{fig:tanh_kr}
  \end{figure}

Since this example is two-dimensional, our map parameterization is immediately strictly triangular, and thus we expect \MGAN to approximate the KR rearrangement, as the latter is a global minimizer of~\eqref{abstract-optimization-problem}.
\Cref{fig:tanh_kr} compares the second component function of the true KR rearrangement to the \MGAN map $\G$, with $\lambda = 0.01$. Interestingly,  the \MGAN map approximates the KR map very closely,
despite not using the explicit KR construction.

\subsection{Insensitivity to variable ordering}\label{sec:banana}
We now illustrate a benefit of using non-triangular maps, rather than triangular maps such as the KR rearrangement. Consider the random vector $u = (u_1, u_2)$, with $u_1 \sim \N(0,1)$ and $u_2|u_1 \sim \N(u_1^2 + 1, 0.5^2)$. For simplicity, we omit the conditioning variables $y$ in this example.
The bivariate distribution of $u$ can be represented exactly as the pushforward of $\eta_\tU = \N(0, I_2)$ by the map $\T(v) = [v_1; v_1^2 + 1 + 0.5 v_2]$. 
Hence, $\T$ is easily approximated by a triangular map of this form.
When the ordering of $u_1$ and $u_2$ are reversed, however---i.e., when the first component of $\T$ must represent the marginal of $u_2$ instead of $u_1$---the triangular map is more challenging to approximate.
To resolve this issue, one common approach is to compose many maps to define an expressive normalizing flow; see~\cite{papamakarios2017MAF} for a similar application. We demonstrate instead that by using a non-triangular parameterization (which would become block triangular when there are conditioning variables), we can avoid issues pertaining to the ordering of the $\u$ variables
and achieve a more robust map in practice.

We use $N = 10^4$ training samples, $\lambda = 0.01$, and the \LSGAN loss function to train
a \MGAN with either triangular or non-triangular structure. We use three-layer fully connected neural networks with hidden layer 
sizes 32/64/32 for the non-triangular maps and neural networks with hidden layer sizes 22/46/22 for each component of the triangular map. In total, the non-triangular and triangular maps have about the same number of parameters. 
 Both \MGAN maps are trained %
 using the same optimization setup
as in \Cref{sec:synthetic-example}.

\begin{figure}[!ht] 
  \centering
  
    \begin{subfigure}[b]{0.25\textwidth}
        \begin{overpic}[width=\linewidth]{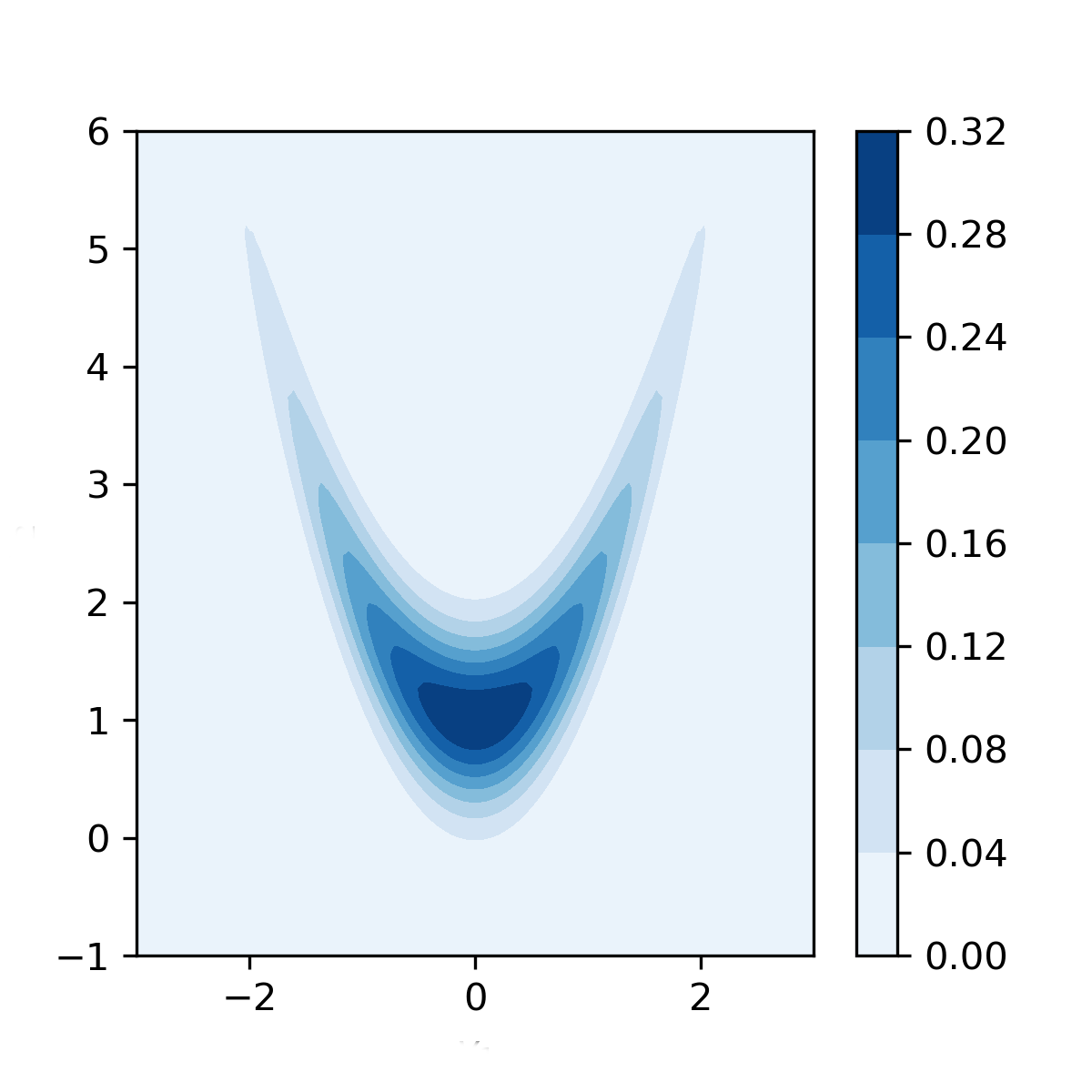}
        \put(0, 50){{\tiny $u_2$}}
        \put(41, 2){{\tiny $u_1$}}
      \end{overpic}
    \end{subfigure}
    \begin{subfigure}[b]{0.25\textwidth}
        \begin{overpic}[width=\linewidth]{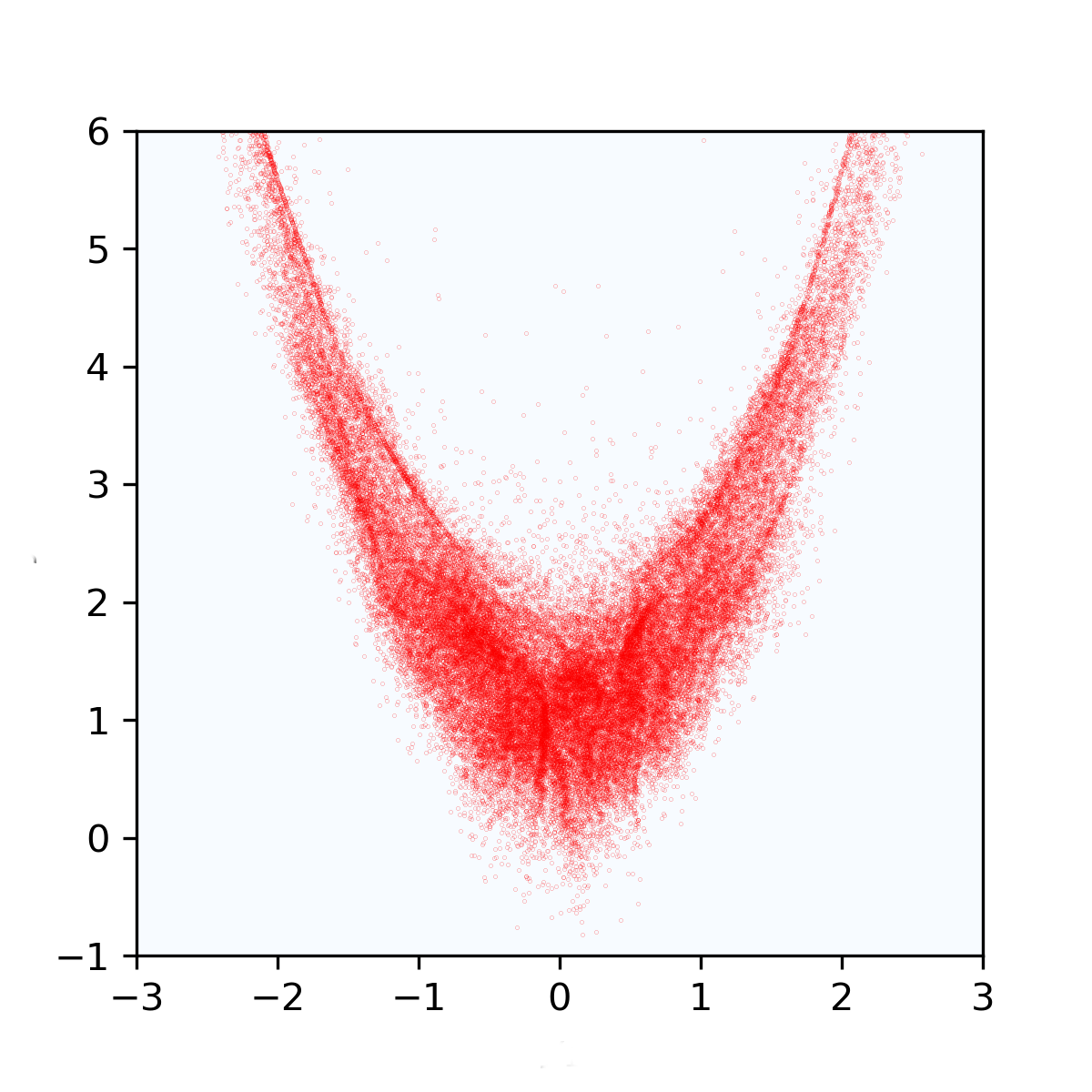}
        \put(50, 2){{\tiny $u_1$}}
      \end{overpic}
      \end{subfigure}
    \begin{subfigure}[b]{0.25\textwidth}
        \begin{overpic}[width=\linewidth]{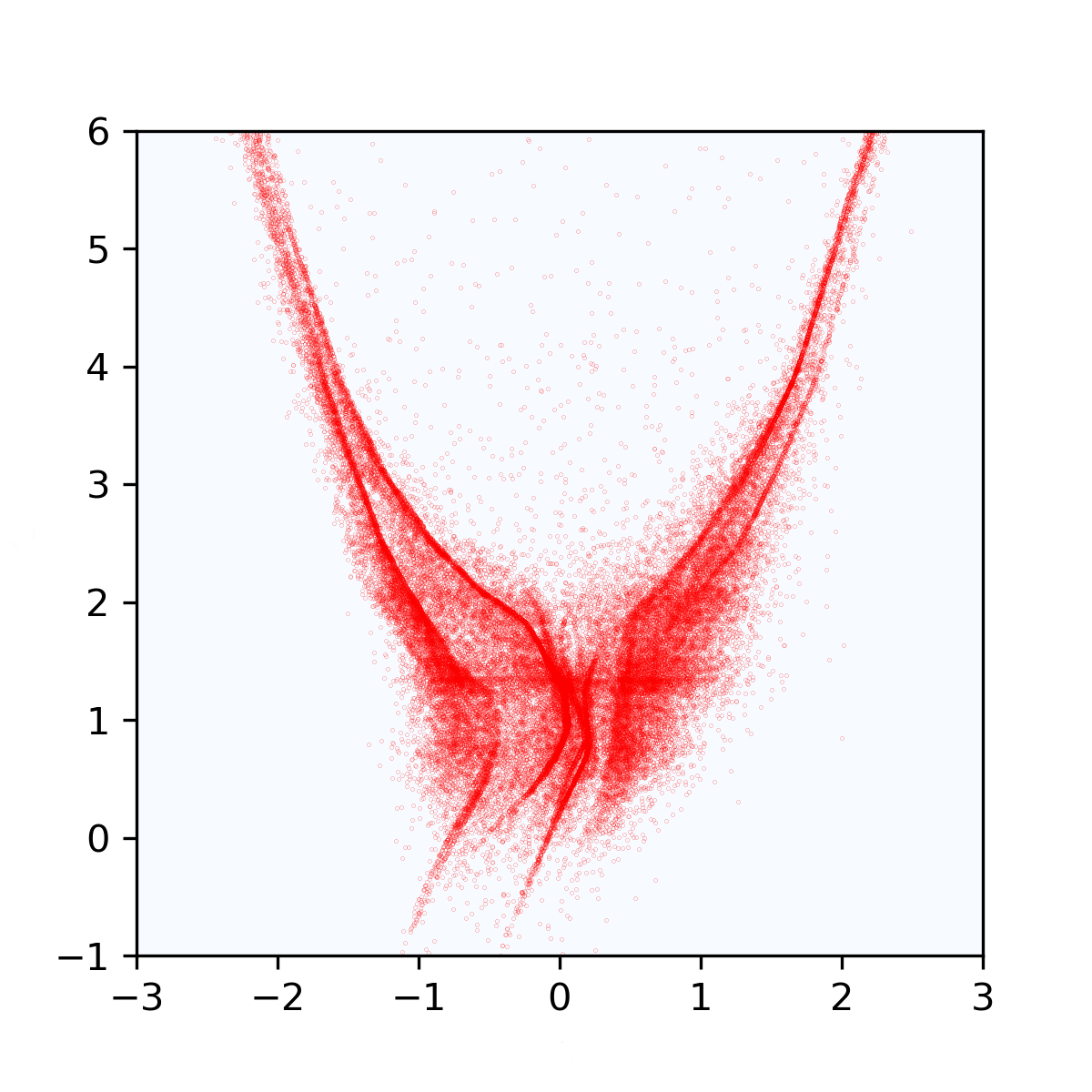}
        \put(50, 2){{\tiny $u_1$}}
      \end{overpic}
      \end{subfigure}
      \caption{(Left) The true density of $(u_1, u_2)$ considered in \Cref{sec:banana}. (Middle) 
        Samples generated by \MGAN using a non-triangular map with the reverse ordering of
        the variables. (Right) Samples generated by \MGAN using a triangular map, also with
      reverse ordering.}
    \label{fig:banana-density} 
  \end{figure}

\Cref{fig:banana-density} compares the samples generated by the triangular and
non-triangular maps to the true density of $u$. We observe that the non-triangular map
is
able to capture the target density with an unfavorable ordering of the variables,
unlike the triangular map. \Cref{tab:banana-KL} reports the KL divergence between the true and approximated distributions for both variable orderings. The non-triangular map provides essentially the same performance independent of ordering, while the performance of the triangular map improves or degrades significantly depending on the ordering. This suggests that non-triangular maps are less sensitive to the variable ordering, a 
major advantage of \MGANs in comparison to autoregressive models where it is necessary to specify a variable ordering in advance.
  
\begin{table}[!ht]
\begin{center}
\begin{tabular}{ c|c|c} 
  \hline
  & Non-triangular & Triangular \\ \hline
  Favorable order $(u_1,u_2)$ & 0.056 (0.003) & 0.039 (0.002) \\ \hline
  Reverse order $(u_2,u_1)$ & 0.058 (0.002) & 0.102 (0.004) \\ \hline
\end{tabular}
\end{center}
\caption{KL divergence errors for non-triangular and triangular \MGANs computed using $N = 10^4$ training samples. The approximate densities are estimated using KDE by generating $5\times 10^4$ samples and using an optimal bandwidth parameter that is chosen using 5-fold cross-validation. The KL divergence is evaluated using an average of $10^4$ independent test samples and is reported with its 95\% standard error in brackets.~\label{tab:banana-KL}}
\end{table}

\subsection{Approximation of OT maps} \label{sec:numerics_OT}
Now we show how the \MGAN framework using an average monotonicity penalty recovers the  transport map $\G$ that minimizes the $L^2$ transport cost $\mathbb{E}_{(y,v) \sim \eta}\|v - \G(y,v)\|^2$, i.e., the conditional Brenier map of \Cref{prop:opt_maps}. 
We consider a multivariate Gaussian distribution $\nu$ with $\Y = \mathbb{R}$ and $\U = \mathbb{R}^5$. The marginal distribution of $\u$ is chosen to be $\mathcal{N}(m_\u,\Sigma_\u)$ where the mean and covariance are randomly sampled as $m_\u \sim \mathcal{N}(0,\text{I}_5)$ and $\Sigma_\u = U U^\top$ for orthonormal column vectors $U \in \R^{5 \times 5}$ (from the QR decomposition of a matrix with standard Gaussian entries) and fixed for this experiment. 
The measurement $\y$ is given by $\y = \u_4 + \xi$ where $\xi \sim \mathcal{N}(0,1)$. By \Cref{prop:opt_maps}, the monotone transport map pushing forward a standard Gaussian reference $\eta_{\tU} = \N(0, \text{I}_5)$ to the conditionals $\nu (\cdot | y)$ is unique among all maps that are gradients of a convex function. In this Gaussian case, the optimal map is given by $\G^\dagger(y,v) \coloneqq m_{u|y}(y) + \Sigma_{\u|y}^{1/2} v$; see \cite[Example~2.1]{carlier2016vector}. 

Given $N = 10^4$ training samples from $\nu$, we learn the \MGAN map $\G$ using the \WGANGP loss with gradient penalty $\gamma = 1$ and three increasing values of the monotonicity penalty $\lambda$. 
We parameterize the maps and the discriminators using three-layer, fully connected neural networks with hidden layer sizes 64/64/64. While affine maps (in $v$ and $y$) are sufficient to represent the Gaussian conditionals in this example, our goal is to demonstrate the convergence of the \MGAN training procedure to the conditional Brenier map over the large space of nonlinear functions described in \Cref{subsec:NN-discretization}. We train using the Adam algorithm as in \Cref{sec:synthetic-example}, with a batch size of $M=1000$ and a scheduled learning rate that decays by $0.995$ starting from $4 \times 10^{-3}$, over $1000$ epochs.

\begin{figure}[!ht]
    \centering
    \begin{subfigure}[b]{0.35\textwidth}
      \includegraphics[width=\textwidth]{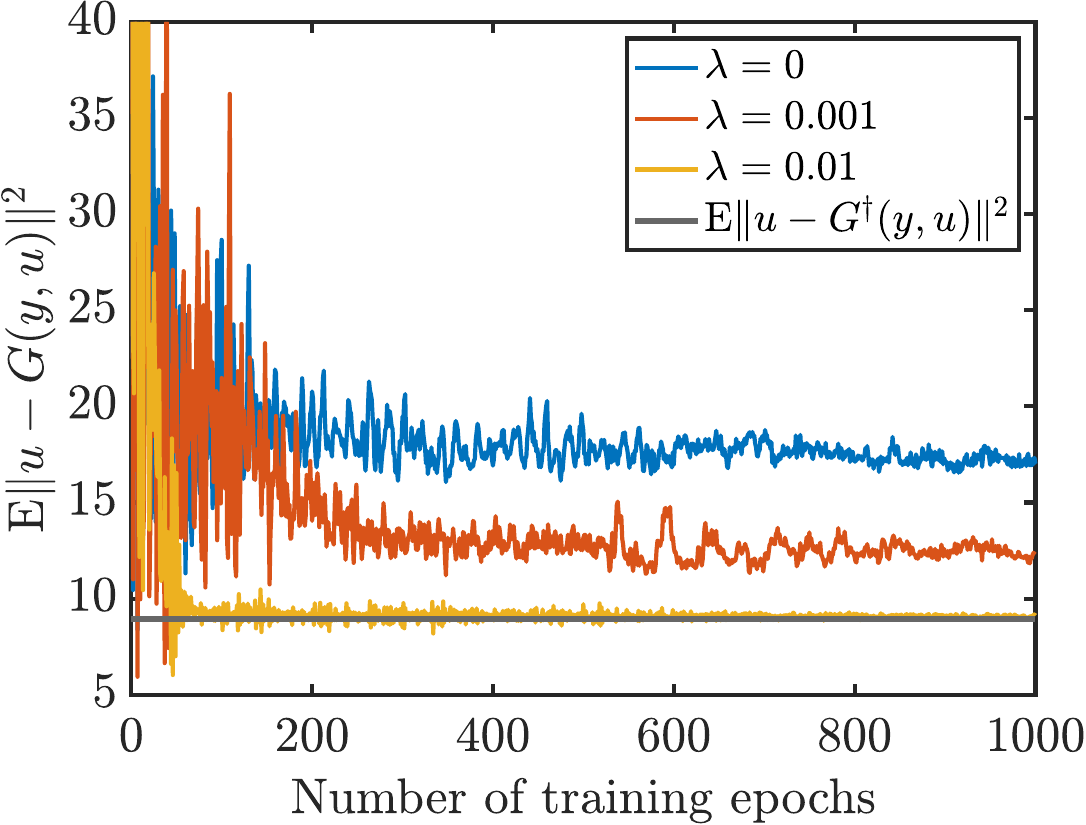}
    \end{subfigure}
    \qquad
    \begin{subfigure}[b]{0.35\textwidth}
      \includegraphics[width=\textwidth]{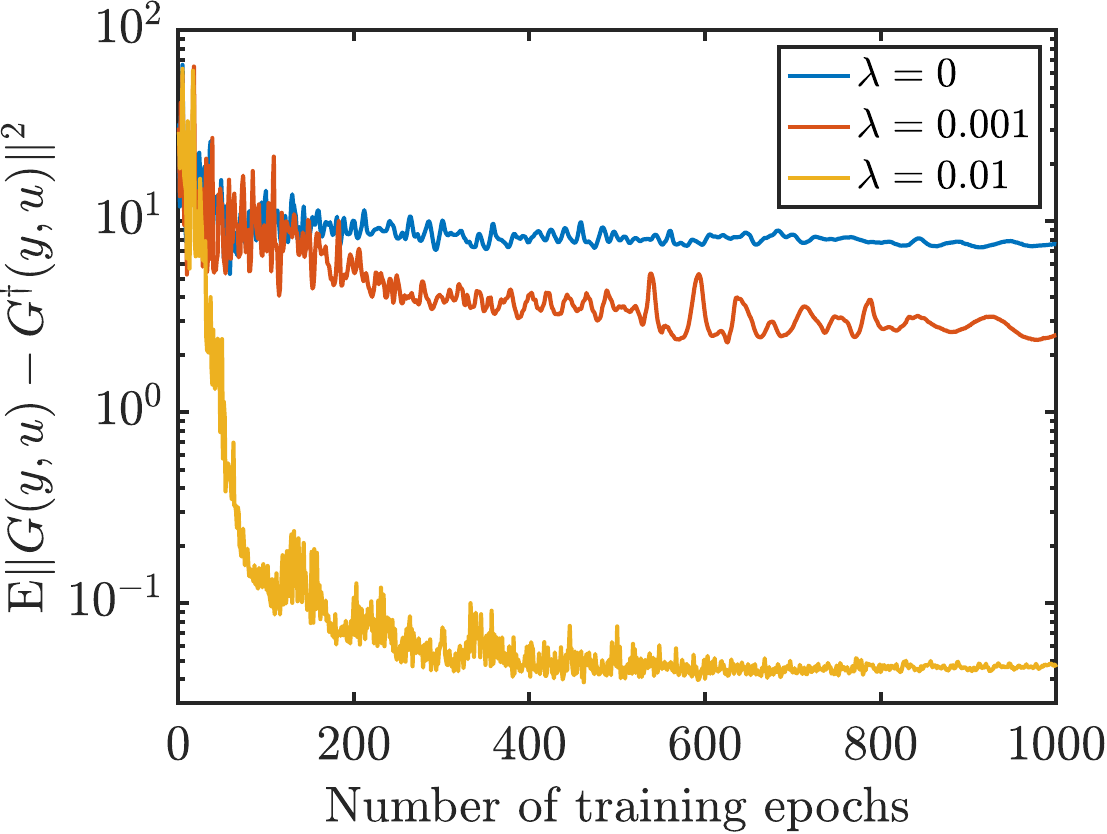}
    \end{subfigure}
    \caption{(Left) The transport cost $\mathbb{E}_{(y,v) \sim \eta}\|v - \G(\y,v)\|^2$ associated with \MGAN maps $\G$ converges to the \emph{minimal} transport cost, achieved by the optimal map $\G^\dagger$, when increasing the monotonicity penalty $\lambda$. (Right) The maps themselves converge in the $L^2_\eta$ sense to the optimal map $\G^\dagger$ when increasing $\lambda$. \label{fig:optimal_map}}
  \end{figure}

\Cref{fig:optimal_map} plots the transport cost $\mathbb{E}_{(y,v) \sim \eta}\|v - \G(y,v)\|^2$ for the estimated maps $\G$ and the expected squared error between the estimated maps and the optimal map $\G^\dagger$. We observe that maps found with the average monotonicity penalty term indeed converge to the \emph{optimal map} of \Cref{prop:opt_maps} when increasing the penalty $\lambda$, and similarly that the associated transport cost converges to the OT cost. Of course, an alternative approach could involve replacing the $\lambda$-dependent monotonicity penalty with the constraint that $\T$ lie in $\mathcal{T}^B$; this involves some practical difficulties, as described in \Cref{remark:icnns}. One 
simple approach in the Gaussian case would be to write 
$\T$ as the gradient of a quadratic function as in
\cite{taghvaei2022optimal, al2023optimal}.

\subsection{Inference of ODE parameters} \label{sec:ODE-inverse}
\begin{figure}[!htb]
  \centering
    \begin{subfigure}[b]{0.4\textwidth}
        \includegraphics[width=\linewidth]{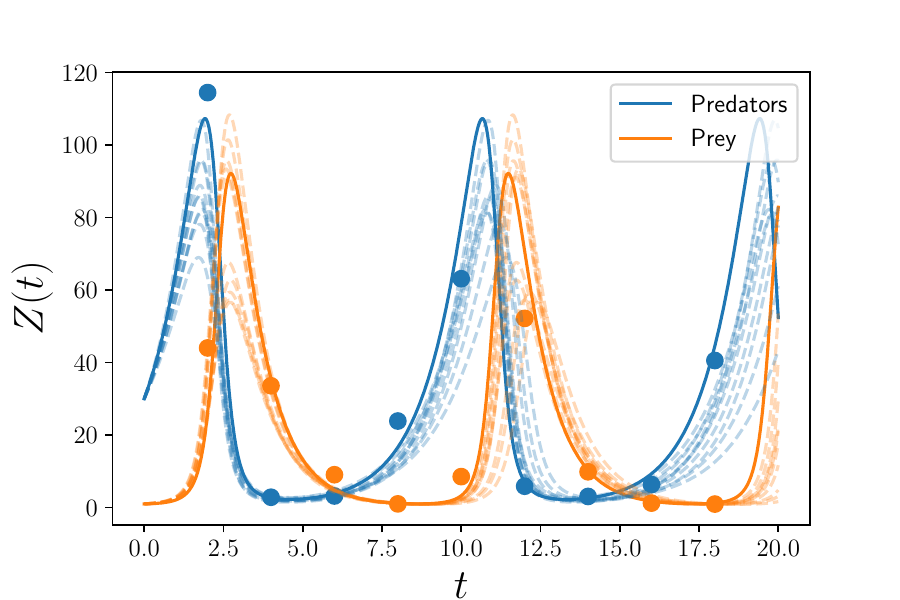}
    \end{subfigure}
    \qquad
    \begin{subfigure}[b]{0.4\textwidth}
        \includegraphics[width=\linewidth]{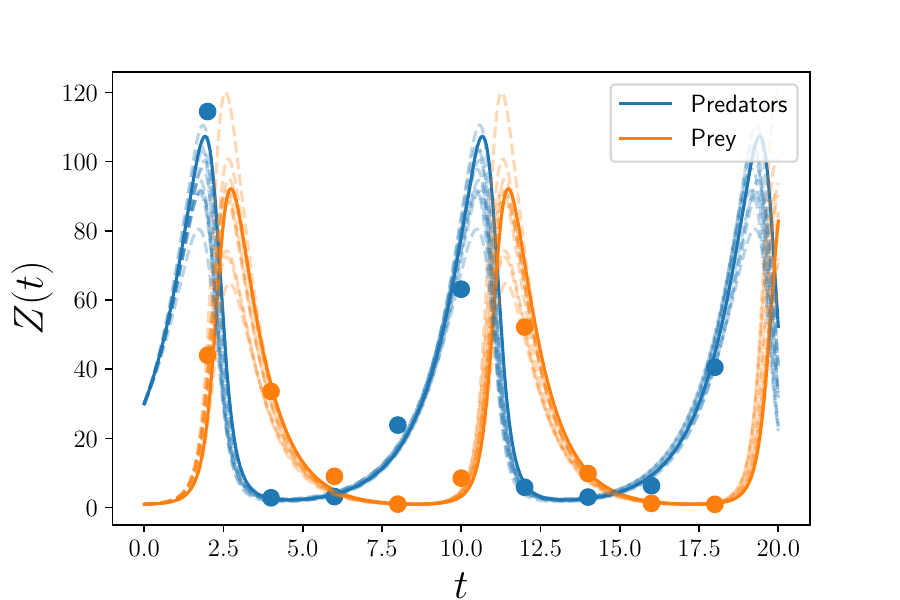}
    \end{subfigure}
    \vspace{-0.2cm}
    \caption{Posterior predictive samples for the states $p(t)$ given ten posterior samples from the (Left) \MGAN map and the (Right) MCMC algorithm.~\label{fig:ODE_postpredictive}}
\end{figure}

Next, we use the MGAN framework to infer the parameters in a Lotka--Volterra population model, which is a common benchmark for likelihood-free inference~\cite{lueckmann2021benchmarking}. This model describes the populations of interacting species, such as predators and prey, using nonlinear coupled ODEs where the rates of change of the two populations depend on four parameters $\u = (\alpha,\beta,\gamma,\delta) \in \R^4$. Our goal is to infer these parameters given noisy observations of the populations of predators and prey, i.e., the system states, at select times. %
The states $p(t) \in \RR_{+}^2$ evolve according to the coupled ODEs
\begin{align*}
\frac{d p_1}{d t} &= \alpha p_1(t) - \beta p_1(t) p_2(t), \\
\frac{d p_2}{d t} &= -\gamma p_2(t) + \delta p_1(t) p_2(t),
\end{align*}
with the initial condition $p(0) = (30,1)$. We simulate the ODEs for $T = 20$ time units and collect noisy observations of the state every $\Delta t_{\text{obs}} = 2$ time units. The  observations are corrupted with log-normal noise, i.e., $\log\y_k \sim \mathcal{N}( \log p(k\Delta t_{\text{obs}}), \sigma^2 \text{I}_2)$ for $k=1,\dots,9$, with standard deviation $\sigma = 0.01$. For inference, we use an independent log-normal prior distribution for the parameters given by $\log \u \sim \mathcal{N}(m_\u,0.5 \text{I}_4)$ with $m_u = (-0.125,  -3, -0.125, -3)$. \Cref{fig:ODE_postpredictive} displays the states $p(t)$ (solid line) for the parameter $\u^\ast = (0.92, 0.05, 1.50, 0.02)$ and an observation $\y^\ast \in \RR^{18}$ drawn from the conditional distribution $\nu(\cdot|\u^*)$.

We then sample from the posterior density for $\u|\y = \y^\ast$ given $N=10^5$ training samples from $\nu$ using both \MGAN and a MCMC algorithm. First, we train an \MGAN network with the \WGANGP loss using the monotonicity penalty $\lambda = 0.1$ and the gradient penalty $\gamma = 1$. For this example we used three-layer, fully-connected neural networks with hidden layer sizes \(128/256/512\) for the map and hidden layer sizes \(512/256/128\) for the discriminator. We used the Adam optimizer with the same parameters as in \Cref{sec:synthetic-example} and trained  for 400 epochs.

\begin{figure}[!htb]
  \centering
    \begin{subfigure}[b]{0.45\textwidth}
        \includegraphics[width=\linewidth]{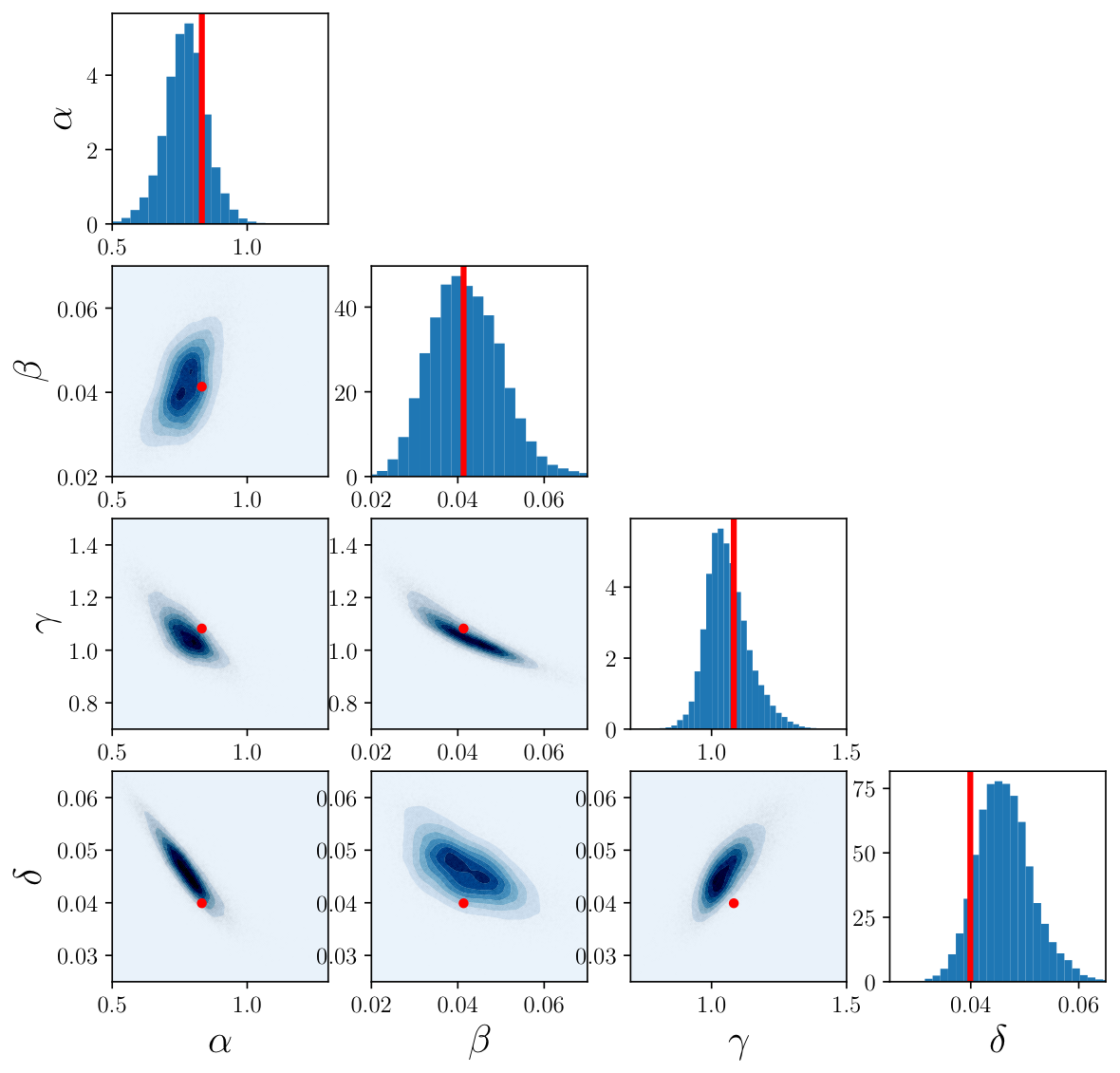}
    \end{subfigure}
    \hfill
    \begin{subfigure}[b]{0.45\textwidth}
        \includegraphics[width=\linewidth]{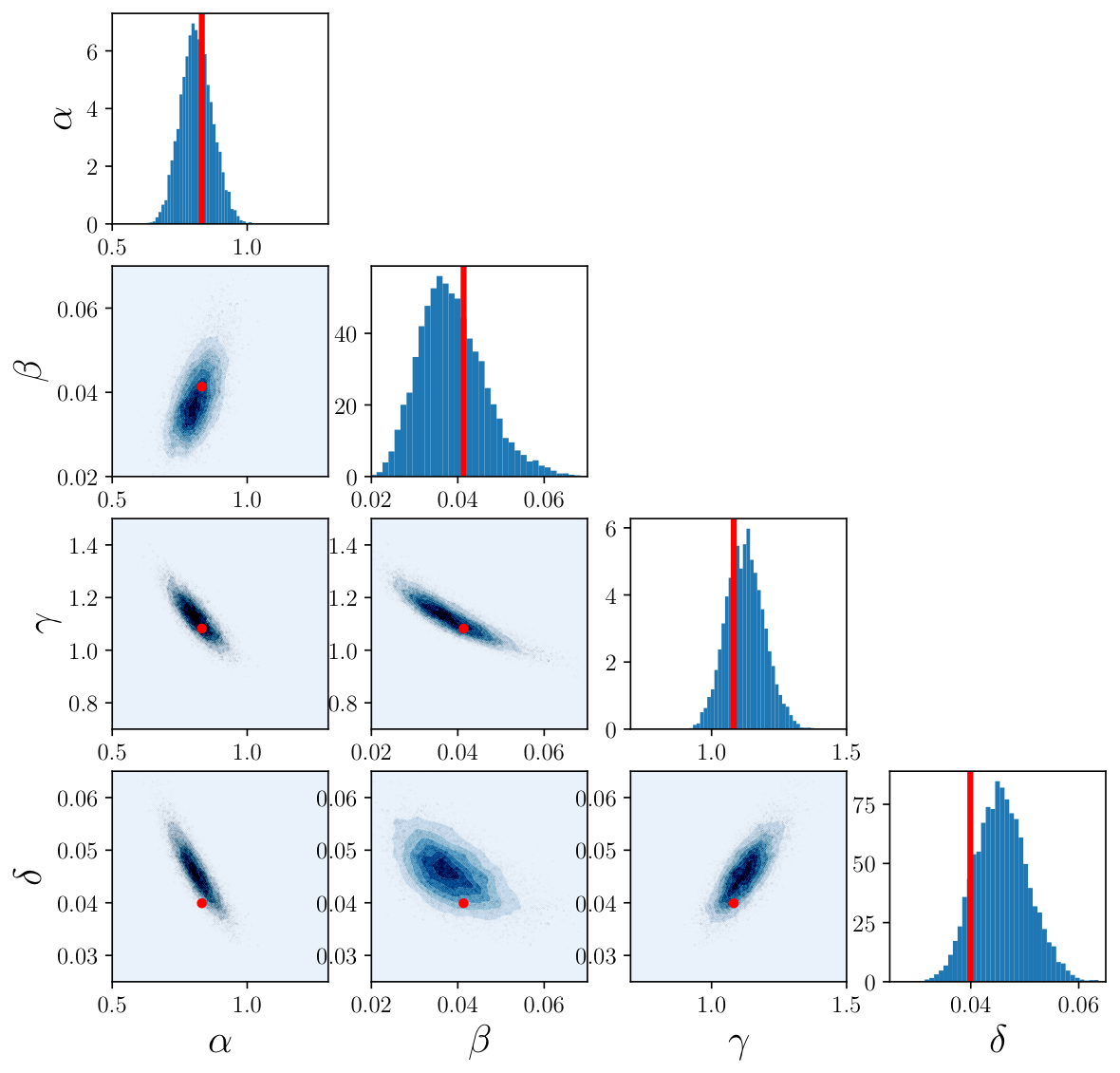}
    \end{subfigure}
    \vspace{-0.2cm}
    \caption{Posterior samples of the parameters in the deterministic Lotka--Volterra model using (left) the \MGAN framework, and (right) the adaptive Metropolis MCMC algorithm. \label{fig:deterministicLV_results}}
\end{figure}

\Cref{fig:deterministicLV_results} displays $100,000$ parameter samples from \MGAN, i.e., $\G(\y^\ast,\u^i)$ for $\u^i \sim \mathcal{N}(0,\text{I}_4)$ after learning the map $\G$, and from an adaptive Metropolis MCMC sampler, respectively. We observe similar one and two-dimensional marginal distributions using both methods. The true parameter $\u^\ast$ that generated the data (denoted in red) is contained in the bulk of the posterior distributions, and appears like a representative sample. Lastly, we integrate the ODEs for sample realizations of the posterior parameters to sample from the predictive distribution for the states $p(t)$. The dashed lines in \Cref{fig:ODE_postpredictive} plot ten posterior predictive samples for both \MGAN and MCMC. We observe that samples from both methods concentrate around the true states and that the predictions from \MGAN have similar spread to  MCMC (i.e., the ground truth), especially at earlier times.

\subsection{Darcy flow Bayesian inverse problem}\label{sec:darcy-flow}

\begin{figure}[htp] 
    \centering
    \begin{subfigure}[b]{0.28\textwidth}
        \includegraphics[width=0.95\linewidth]{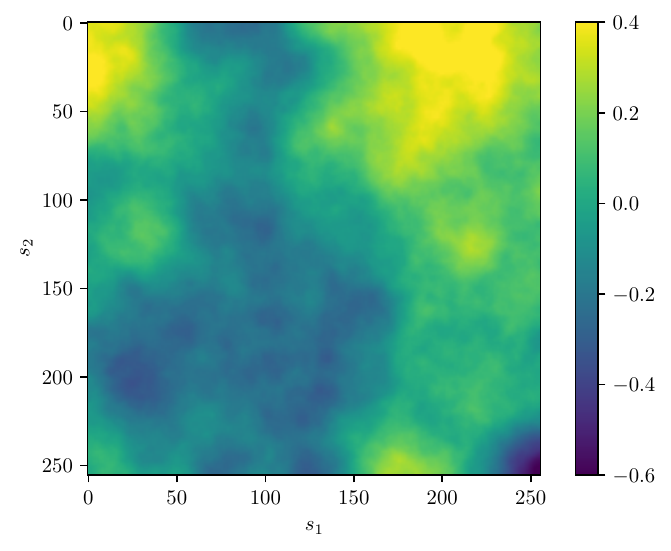}
        \caption*{(a) True field $u$}
      \end{subfigure}
    \begin{subfigure}[b]{0.28\textwidth}
        \includegraphics[width=0.95\linewidth]{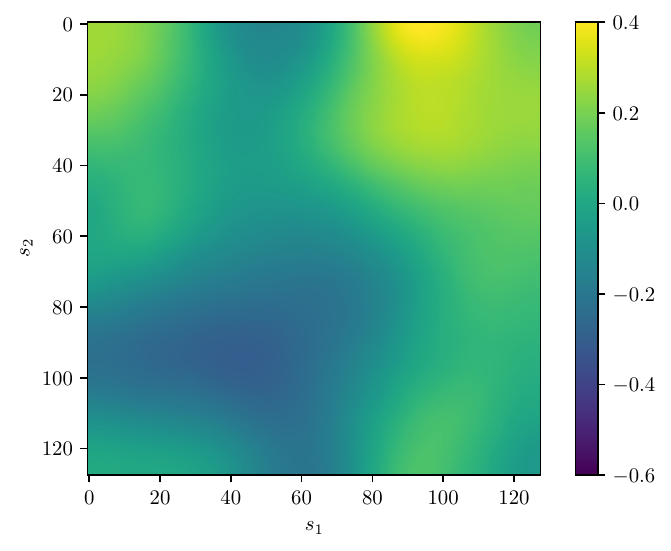}
        \caption*{(c) pCN mean}
      \end{subfigure}
    \begin{subfigure}[b]{0.28\textwidth}
        \includegraphics[width=0.95\linewidth]{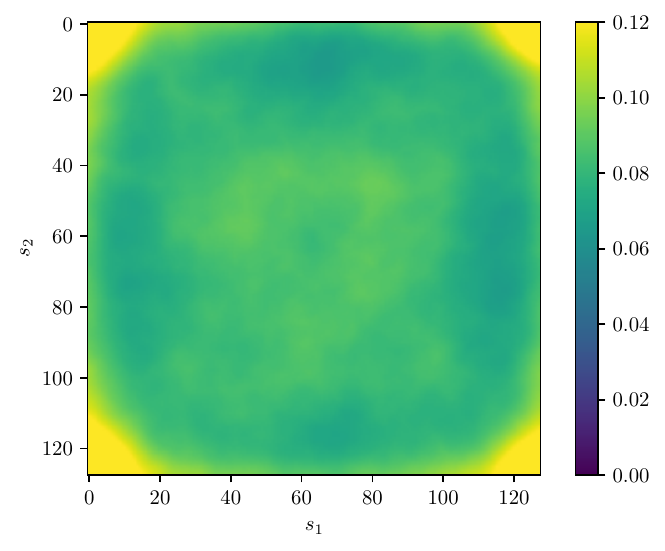}
        \caption*{(e) pCN std}
    \end{subfigure}\\
    \begin{subfigure}[b]{0.28\textwidth}
        \includegraphics[width=0.95\linewidth]{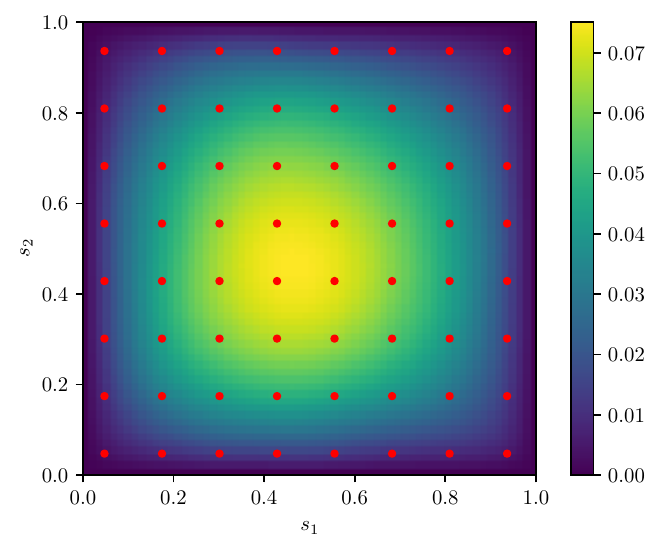}
        \caption*{(b) Pressure field $p$}
    \end{subfigure}
    \begin{subfigure}[b]{0.28\textwidth}
        \includegraphics[width=0.95\linewidth]{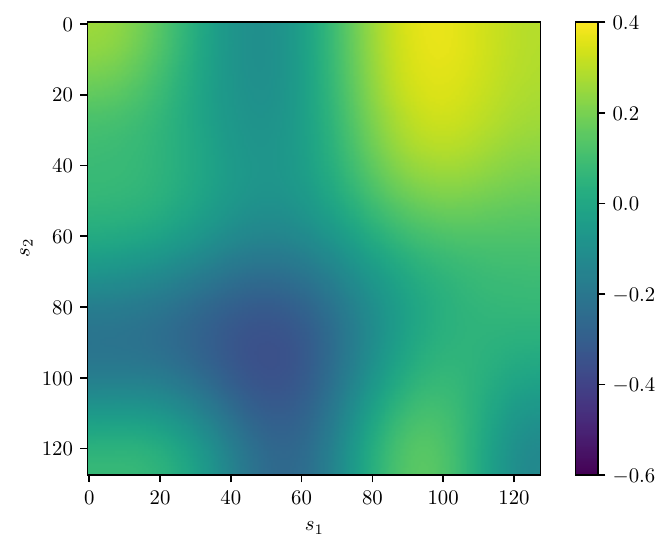}
        \caption*{(d) \MGAN mean}
    \end{subfigure}
    \begin{subfigure}[b]{0.28\textwidth}
        \includegraphics[width=0.95\linewidth]{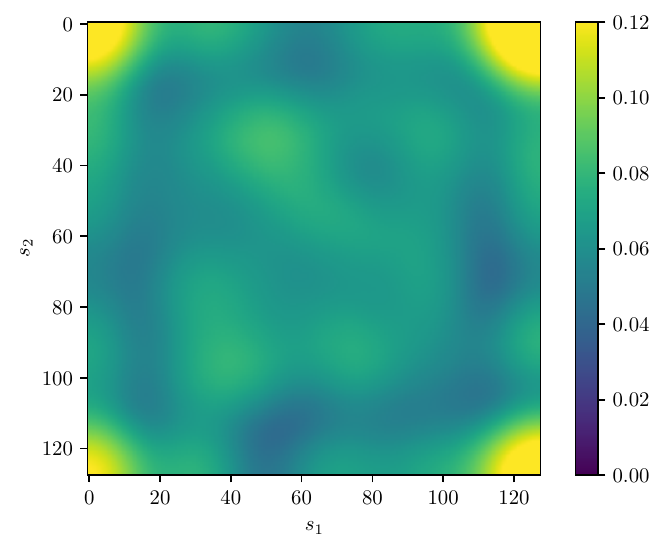}
        \caption*{(f) \MGAN std}
    \end{subfigure}

    \caption{A comparison of posterior samples for the Darcy flow inverse problem using \MGANs and pCN.
      (a) A realization of the Gaussian random field $u$. (b) The solution to the PDE with the measurement locations denoted in red. (c) and (d) The mean of the posterior samples from \MGAN and pCN, respectively. (e) and (f) The standard deviation of the posterior samples from \MGAN and pCN, respectively.~\label{fig:DarcyFlow-fine}}
\end{figure}

We now consider a benchmark inverse problem  from
subsurface flow modeling \cite{iglesias-subsurface} and electrical impedance tomography \cite{somersalo} whose forward model is given by the partial differential equation (PDE)
\begin{equation}
\label{eq:darcy-flow}
\begin{aligned}
- \nabla \cdot (a(s) \nabla p(s)) &= 1, && s \in (0,1)^2, \\
p(s) &= 0, && s \in \partial (0,1)^2.
\end{aligned}
\end{equation}
We interpret \(p(s)\) as the pressure field of subsurface flow in a reservoir 
with permeability coefficient \(a(s) \in \RR_{+}\) under constant forcing. We further introduce a log-normal random field for the permeability given by $u = \log(a) \sim \mathcal{N}(0,(-\Delta + 9I)^{-2})$ where $\Delta$ is the Laplacian operator with zero Neumann boundary conditions. The inverse problem is to recover the random field $\u$ given noisy measurements $\y$ of the pressure at $64$ regularly spaced locations, i.e., $\y = (p(s_1),\dots,p(s_{64})) + \mb{\gamma}$ where $\mb{\gamma} \sim \mathcal{N}(0,10^{-6}I_{64})$. \Cref{fig:DarcyFlow-fine}(a,d) plot a realization of $u$ along with the solution to the PDE on a grid of size $256 \times 256$
as well as the location of pressure measurements $s_1, \dots, s_{64}$.

To recover the permeability $a$, we train a \MGAN following the discussion of \Cref{subsec:char-BIP}.
We sample a training set of size $N = 10^5$ from $\nu$ that was obtained using the following recipe: draw $u \sim \mathcal{N}(0,(-\Delta + 9I)^{-2})$ and set $a = \exp(u)$; solve the PDE using finite differences to obtain $p$; use spline interpolation
to simulate a set of measurements $\y$ at the observation locations.
The \MGAN map was trained using the  \WGANGP loss functional with the gradient penalty $\gamma = 1$
and the monotonicity penalty $\lambda = 0.01$. To make our network architectures consistent in the continuum limit, and hence mesh independent,
we use PCA projections to reduce the dimension of the $u$ samples at the input to the networks, i.e., the $\Psi^I$ operator in \Cref{subsec:NN-discretization}
is taken to be the PCA projection of the random field $u$ onto its first $25$ PCA modes, which capture 99.98\% of the total prior variation (as measured by the trace of the prior covariance). 
To this end, the $\G$ component of our \MGAN map takes inputs in
$\RR^{64} \times \RR^{25}$ (64 for $\y$ and $25$ for the leading PCA modes of $u$) and outputs a vector of PCA modes of $u$ in $\RR^{25}$ which can then
be lifted to a random field by taking $\Psi^O$ to be the PCA reconstruction map. In summary, our map $\G$ will condition the
first 25 PCA coefficients of $u$ on observations of the data $\y$. In this experiment we use the same network architectures and optimizer as in \Cref{sec:ODE-inverse} and train for 500 epochs.

We used $5 \times 10^4$ prior samples in order to compute the PCA modes of $u$ and generated a fixed realization $y^\ast$ for a single draw of the field $u$, that is taken to be the ground truth. To avoid any inverse crimes \cite{kaipio2006statistical} we generated the data $y^*$ using a mesh
that was twice as fine as the mesh used to generate the training data. \Cref{fig:DarcyFlow-fine}(b,e,c,f) compares
the posterior mean and the standard deviation for the field $u$ obtained by \MGANs with the preconditioned Crank-Nicolson (pCN) MCMC algorithm~\cite{stuart-mcmc}, which is 
regarded as the gold standard solution. We tuned the pCN step size to achieve an acceptance rate between $20\%$ and $40\%$ after burn-in and used $10^6$ samples
to compute the mean and standard deviations. We observe good agreement between the \MGAN mean and pCN while the standard deviation appears to have been
slightly under estimated by \MGAN, a feature that is common with prior-based dimension reduction techniques. Note that a more conservative estimate for the standard deviation can be obtained by sampling the trailing PCA coefficients from their prior distribution, i.e., without conditioning on the data, and combining these with the \MGAN posterior samples for the leading PCA coefficients \cite{bigoni2019greedy,cui2014likelihood}.

This experiment not only demonstrates the feasibility of \MGANs for likelihood-free inference on function spaces, but also suggests 
that \MGANs can potentially lead to improved performance for 
PDE inverse problems:
Our maps were computed using only $10^5$ PDE solves while pCN required $10^6$ samples to compute a stable estimate of the standard deviation. Moreover, the latter would have to be re-run for any new realization of the data $y$, whereas the \MGAN map can be applied, without additional training, to any new realization of $y$.
Furthermore, the \MGAN training set can be generated fully in parallel since its samples are independent, unlike MCMC that requires sequential PDE solves
for each accept/reject step.

\subsection{Probabilistic image in-painting}\label{sec:image-painting}

For our final set of experiments, we consider the ``in-painting'' problem of reconstructing an image after a portion of it has been removed. 
 We view this problem in our general probabilistic setting, as image-to-image regression where the input/measurement $\y$ is the incomplete image and the output $u$ is the in-painting. The conditional distribution $\nu( \cdot | y)$ thus quantifies uncertainty in the reconstruction, with its samples being understood as candidate in-paintings.
We consider the CelebA dataset consisting of \( 64 \times 64 \times 3 \)
RGB images of celebrity faces (converted to a standard size using
bi-cubic interpolation). The input $\y \in \RR^{32 \times 64 \times 3}$ consists of the top half of each image,
and the output \( \u \in \RR^{32 \times 64 \times 3}\) consists of the bottom half.

We trained an \MGAN on the training set of $N = 162\,770$ images using the \WGANGP loss functional 
with the monotonicity penalty $\lambda = 10^{-4}$ and  gradient penalty $\gamma = 1$. 
We also added independent Gaussian white noise with standard deviation $0.05$ to corrupt each image 
in the training set, as in~\cite{karras2019style}.
We chose our reference measure to be $\eta = \nu_\Y \otimes \N(0, \text{I}_{100})$, i.e., the $\Y$ marginal coincides with that of the training
data, while the latent space for the $\U$ variable is assumed to be $\RR^{100}$ equipped with standard Gaussian measure. Hence, the input and output spaces of $\T$ do not match in this example, in contrast with our previous experiments. As for the architectures, we used 
 the convolutional architectures introduced in \cite{pathak2016context} with suitable modifications for our input and output dimensions.
We used the same training/optimization setup as in \Cref{sec:synthetic-example}.

\begin{figure}[!ht] 
  \centering
    \begin{subfigure}[b]{0.10\linewidth}
        \includegraphics[width=\linewidth]{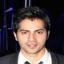}
        \includegraphics[width=\linewidth]{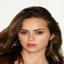}
        \includegraphics[width=\linewidth]{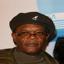}
        \begin{overpic}[width=\linewidth]{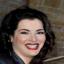}
          \put(18,-18){\scriptsize Truth}
        \end{overpic}
    \end{subfigure}
    \begin{subfigure}[b]{0.10\linewidth}
        \includegraphics[width=\linewidth]{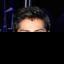}
        \includegraphics[width=\linewidth]{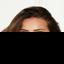}
        \includegraphics[width=\linewidth]{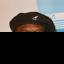}
        \begin{overpic}[width=\linewidth]{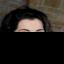}
          \put(35,-18){\scriptsize $\y^\ast$}
        \end{overpic}
    \end{subfigure}
    \begin{subfigure}[b]{0.10\linewidth}
        \includegraphics[width=\linewidth]{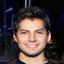}
        \includegraphics[width=\linewidth]{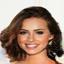}
        \includegraphics[width=\linewidth]{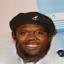}
        \includegraphics[width=\linewidth]{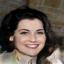}
    \end{subfigure}
    \begin{subfigure}[b]{0.10\linewidth}
        \includegraphics[width=\linewidth]{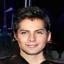}
        \includegraphics[width=\linewidth]{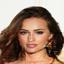}
        \includegraphics[width=\linewidth]{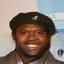}
        \begin{overpic}[width=\linewidth]{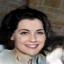}
          \put(45,-18){\scriptsize $u|\y^\ast$ samples}
        \end{overpic}
    \end{subfigure}      
    \begin{subfigure}[b]{0.10\linewidth}
        \includegraphics[width=\linewidth]{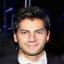}
        \includegraphics[width=\linewidth]{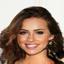}
        \includegraphics[width=\linewidth]{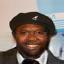}
        \includegraphics[width=\linewidth]{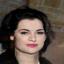}
    \end{subfigure}
    \begin{subfigure}[b]{0.10\textwidth}
        \includegraphics[width=\textwidth]{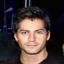}
        \includegraphics[width=\textwidth]{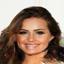}
        \includegraphics[width=\textwidth]{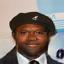}
        \includegraphics[width=\textwidth]{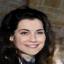}
    \end{subfigure}
    \begin{subfigure}[b]{0.10\linewidth}
        \includegraphics[width=\linewidth]{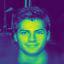}
        \includegraphics[width=\linewidth]{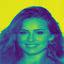}
        \includegraphics[width=\linewidth]{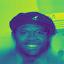}
        \begin{overpic}[width=\linewidth]{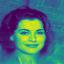}
          \put(18,-18){\scriptsize $\mathbb{E}[u|\y^\ast \!]$}
        \end{overpic}
     \end{subfigure}
    \begin{subfigure}[b]{0.10\linewidth}
        \includegraphics[width=\linewidth]{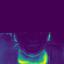}
        \includegraphics[width=\linewidth]{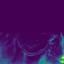}
        \includegraphics[width=\linewidth]{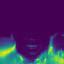}
        \begin{overpic}[width=\linewidth]{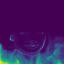}
          \put(18,-18){\scriptsize $\mathbb{V}[u|\y^\ast \!]$}
        \end{overpic}
    \end{subfigure}
    \vspace{0.2cm}
    \caption{Example in-paintings using \MGAN for the CelebA test set. The first column depicts the 
    ground truth image, the second column shows the observed image \(\y^\ast\) while the next three columns
    are random samples from the conditional distribution for \(\u |\y^\ast\). The last two columns
    show the pointwise means and variances for the intensities of the conditional samples generated by the \MGAN map.}
  \label{fig:celebA}
  \end{figure}

 \Cref{fig:celebA} shows conditional samples
 of image in-paintings for the CelebA test set, together with the  conditional mean and variance %
 of the pixelwise image intensities. 
 We note the variability amongst the \MGAN samples,
 producing different smiles, hair styles, jawlines, outfits, and backgrounds---as one 
 should expect from a probabilistic in-painting method. We also computed a FID score of approximately $35$ for
 the \MGAN map in this example. We emphasize, however, that while FID is a common metric for 
 photorealism, it fails to capture accuracy in characterizing the conditional distributions. For example, we noticed that maps
 whose range collapses conditionally onto a single point and as result sample the same in-painting $\G(\y,w)$ for any realization of $w \sim \N(0, I_{100})$ 
 can still obtain
 similarly good FID scores, while failing to capture the true conditional distribution.
To the best of our knowledge, distributional in-painting on the CelebA dataset has not been 
explored in the literature, and thus we cannot compare the FID of our result to others.
The closest to the state of the art is a FID of 30 reported in \cite{lindgren2020conditional} for the CelebA-HQ dataset.

\section{Conclusions}\label{sec:conclusions}

We have developed \MGAN, a transport-based approach for conditional generative modeling and likelihood-free (simulation-based) inference. Our approach seeks a block triangular transport map that pushes forward a chosen reference measure $\eta$ to a target measure $\nu$, defined on the joint space of the parameter and data.
Under very mild assumptions, essentially that the reference measure has an appropriate product structure, we show that this construction produces a component transport map that captures the conditional measures $\nu (\cdot \vert y)$ of the target, and that this map enables direct conditional sampling. We propose an adversarial training procedure to learn such a map, incorporating a monotonicity penalty that drives the solution of the optimization problem towards the unique \textit{conditional} OT map minimizing an $L^2$ transport cost. Numerical experiments demonstrate the effectiveness and versatility of \MGANs in applications ranging from parameter inference and inverse problems to imaging, all tackled in an entirely data-driven/likelihood-free setting.

In future research, the interplay between the quality of an \MGAN map obtained by solving the practical optimization problem~\eqref{practical-approximation-problem} and the accuracy of the derived conditionals warrants theoretical investigation. Relatedly, approximation results characterizing the expressiveness of parametric classes of block triangular maps would be of great interest. It would also be useful to extend the links to OT described here to infinite-dimensional function spaces, as such a connection would be pertinent to inverse problems.

\section*{Acknowledgments}
RB and YM gratefully acknowledge support from the US Department of Energy (DOE) AEOLUS center (award DE-SC0019303) and from the DOE M2dt center (award DE-SC0023187). RB also acknowledges support from an NSERC PGS-D fellowship, the Air Force Office of Scientific Research MURI on ``Machine Learning and Physics-Based Modeling and Simulation'' (award FA9550-20-1-0358), and a Department of Defense (DoD) Vannevar Bush Faculty Fellowship (award N00014-22-1-2790). BH acknowledges support 
from the National Science Foundation research grant 
NSF-DMS-2208535, ``Machine Learning for Bayesian Inverse Problems.'' NBK acknowledges support from the NVIDIA Corporation through full-time employment.

\color{black}

\bibliographystyle{siamplain}
\bibliography{ConditionalSampling_References}

\end{document}